\documentclass[draft]{agujournal2019}
\usepackage{anyfontsize}
\usepackage{url}
\usepackage{xurl}
\usepackage{lineno}
\usepackage{amsmath}
\usepackage[inline]{trackchanges}
\usepackage{soul}
\usepackage{multirow}
\usepackage{diagbox}

\usepackage{ragged2e}                               
\usepackage{enumitem}                               
\setlist{itemsep=0pt, topsep=0pt, parsep=3.5pt}     
\draftfalse

\begin{document}
\justifying 
%
%
\title{Automatic Differentiation-based Full Waveform Inversion with Flexible Workflows}

%
%
\authors{Feng Liu\affil{1,2}, 
        Haipeng Li\affil{2,3*}, 
        Guangyuan Zou\affil{2}, 
        Junlun Li\affil{2,4,5*}}

\affiliation{1}{School of Electronic Information and Electrical Engineering, Shanghai Jiao Tong University, Shanghai 200240, China}
\affiliation{2}{Laboratory of Seismology and Physics of Earth’s Interior, School of Earth and Space Sciences, University of Science and Technology of China, 96 Jinzhai Road, Hefei 230026 Anhui, China.}
\affiliation{3}{Now at Department of Geophysics, Stanford University, CA 94305, USA.}
\affiliation{4}{Mengcheng National Geophysical Observatory, University of Science and Technology of China, Hefei 230026 Anhui, China.}
\affiliation{5}{CAS Center for Excellence in Comparative Planetology, 96 Jinzhai Road, Hefei 230026 Anhui, China.}

\correspondingauthor{Junlun Li}{lijunlun@ustc.edu.cn}
\correspondingauthor{Haipeng Li}{haipeng@sep.stanford.edu}

\begin{keypoints}
\item ADFWI, an automatic differentiation-based, open-source framework, offers an alternative to the adjoint state method in FWI.
\item The framework incorporates various types of wave equations, objective functions and regularization techniques, and allows rapid uncertainty estimation with neural network dropout.
\item With efficient memory management using mini-batches and checkpointing, ADFWI is practical for large-scale full waveform inversion.
\end{keypoints}

%
%
\begin{abstract}

    Full waveform inversion (FWI) is able to construct high-resolution subsurface models by iteratively minimizing discrepancies between observed and simulated seismic data. However, its implementation can be rather involved for complex wave equations, objective functions, or regularization. Recently, automatic differentiation (AD) has proven to be effective in simplifying solutions of various inverse problems, including FWI. In this study, we present an open-source AD-based FWI framework (ADFWI), which is designed to simplify the design, development, and evaluation of novel approaches in FWI with flexibility. The AD-based framework not only includes forword modeling and associated gradient computations for wave equations in various types of media from isotropic acoustic to vertically or horizontally transverse isotropic elastic, but also incorporates a suite of objective functions, regularization techniques, and optimization algorithms. By leveraging state-of-the-art AD, objective functions such as soft dynamic time warping and Wasserstein distance, which are difficult to apply in traditional FWI are also easily integrated into ADFWI. In addition, ADFWI is integrated with deep learning for implicit model reparameterization via neural networks, which not only introduces learned regularization but also allows rapid estimation of uncertainty through dropout. To manage high memory demands in large-scale inversion associated with AD, the proposed framework adopts strategies such as mini-batch and checkpointing. Through comprehensive evaluations, we demonstrate the novelty, practicality and robustness of ADFWI, which can be used to address challenges in FWI and as a workbench for prompt experiments and the development of new inversion strategies.

\end{abstract}

\section*{Plain Language Summary}

    Characterization of the Earth's subsurface is crucial for oil exploration, earthquake studies, environmental assessments, etc. Full waveform inversion (FWI) is a technique that helps scientists create detailed images of the subsurface by comparing observed with simulated seismic data and adjusting the Earth model accordingly. This study introduces ADFWI, an open-source framework dedicated to simplifying FWI with automatic differentiation, which is a technique widely used today in machine learning and inverse problems. ADFWI largely simplifies the use of advanced mathematical techniques and numerical implementations, rendering it easier for researchers to develop and evaluate new approaches in FWI to imaging different subsurface formations. The framework supports a wide range of wave equations in different media and optimization methods, and allows comparisons of different strategies efficiently. Integration of ADFWI with deep learning can further improve inversion stability and expedite uncertainty assessments. Overall, comprehensive evaluations show that ADFWI is reliable and user-friendly, and is useful for researchers to tackle challenges in obtaining complex subsurface structures.
    
%
%
\section{Introduction}
    Full waveform inversion (FWI) is a well-established method used to reconstruct high-resolution subsurface velocity and density models through iteratively minimizing the discrepancy between observed and synthetic seismic data \cite{lailly_1983_Seismic, tarantola1984linearized}. Over the past few decades, FWI has proven effective in geophysical challenges in imaging complex structures with accuracy and resolution beyond traditional tomography \cite{virieux_2009_Overview, tromp2020seismic}. In addition to applications in hydrocarbon exploration \cite{Plessix_2013_Multiparameter, warner_2013_Anisotropic}, FWI has also been widely applied to reveal continental, oceanic as well as global subsurface structures with high resolution \cite{operto2006crustal, bleibinhaus2007structure, fichtner2009full, morgan2013next, zhu_2015_seismic, bozdaug2016global, li2023continent}. Recently, time-lapse FWI has been used to monitor subtle subsurface changes \cite{hicks2016time, egorov2017time, nakata2022seismic}. While accurate wavefield simulations are instrumental for FWI, the elastic characteristics of the Earth are often overlooked when using simple wave physics such as acoustic wave equations \cite{virieux_2009_Overview, agudo_2018_acoustic}. With continuous improvement in high-performance computing, elastic FWI is increasingly used in practice to invert for multiple elastic moduli, albeit at the cost of increased computational complexity and cost \cite{Brossier_2009_Seismic, Plessix_2013_Multiparameter}. More realistic subsurface properties can be obtained by anisotropic elastic FWI, which accounts for orientation-dependent wave propagation in fractured reservoirs, layered sediments \cite{alkhalifah_2000_Acoustic, warner_2013_Anisotropic} or upper mantle structures \cite{zhu_2015_seismic, fichtner_2015_crust}. Nevertheless, honoring more realistic physics in seismic wave propagation inevitably increases computational costs and introduces complexity to derivation and numerical implementation in FWI. 
    
    FWI is typically cast as a local optimization problem \cite{tarantola1984linearized}. The optimization involves calculating the gradient of the objective function with respect to model parameters of interest, where the adjoint state method (ASM) is often used for the gradient computation \cite{liu_2006_FiniteFrequency, fichtner_2006_Adjoint, plessix_2006_review}. For self-adjoint wave equations, the same propagator used in forward modeling can also be applied in the adjoint modeling \cite{bube2016self}. However, for non-self-adjoint wave equations, such as the first-order stress-velocity formulations of the elastic wave equations, separate theoretical derivations and numerical implementations are required for the forward and the corresponding discretized adjoint propagators. This process can be time-consuming and error-prone, and often requires careful verification of the adjointness of the implemented forward and adjoint operators with dot-product tests \cite{claerbout_2008_basic}. Especially, when using advanced finite-difference stencils, adopting absorbing and free-surface boundary conditions, or dealing with wave equations in more intricate subsurface media involving inelasticity and anisotropy, the workload for derivation and implementation of the adjoint propagator can be rather demanding. 

    Also, the choice of the objective function is critical for the convergence and resolution of FWI. The conventional L2-norm-based objective function, though accurate, is sensitive to non-Gaussian noise and prone to cycle skipping \cite{brossier_2010_Which}. Alternatives like envelope-based objective functions can reduce dependence on the initial model and provide improved convexity \cite{bozdag_2011_Misfit, wu_2014_seismic}. Additionally, objective functions originated from deep learning, such as the soft dynamic time warping (soft-DTW) and Wasserstein distance with sinkhorn divergence, have improved performance compared to their traditional counterparts \cite{ma_2013_wave, engquist_2016_Optimal} in terms of cycle skipping and efficiency. However, challenges in the FWI implementation are further compounded when different objective functions are considered, since each objective function needs separate derivation of an adjoint source in the ASM framework. Therefore, a unified framework that is applicable to different types of wave equations and objective functions with reduced derivation and implementation complexity is highly desirable.

    Automatic differentiation (AD) is a technique to obtain gradients of complex differentiable functions with respect to multiple parameters, which is commonly referred to as "back-propagation" in the deep learning community \cite{rumelhart_1986_Learning, baydin_2015_Automatic, lecun_2015_Deep, goodfellow_2016_Deep}. Compared to ASM, AD provides an efficient and easier approach to compute derivatives, eliminating the need for manual derivation and implementation of adjoint equations and sources. More specifically, AD can automatically track the gradient of intermediate variables with respect to input parameters during forward modeling. However, this process typically requires dedicated computational tools, which used to introduce practical challenges such as reduced code generality and limited extensibility \cite{sambridge_2007_Automatic, cao_2015_Computational}. Additionally, AD can consume substantial memory when tackling large seismic datasets or complex models \cite{yang_2014_RTM, wang_2023_Memory}. 

    In the past decades, AD has been extensively used in fluid dynamics \cite{rath_2006_Joint, bischof_2007_Automatic, bezgin_2024_JAXFluids}, atmospheric science \cite{carmichael_1997_Sensitivity, kochkov_2024_Neural, xiao_2024_FengWu4DVar} and geophysical inverse problems \cite{sambridge_2007_Automatic, abokhodair_2009_Complex, stanziola_2023_Jwave, liu_2024_Multimodal}. In seismology, previous studies were limited by the available AD tools and mainly focused on simplified inverse problems. For example, \citeA{sambridge_2007_Automatic} use the forward mode in AD to calculate the gradients in ray tracing and receiver functions, and also demonstrates the efficiency of reverse-mode-based gradient calculations in minimizing the Gibbs free energy. Recently, AD tools have been increasingly used in more complicated FWI. \citeA{tan_2010_Verification} attempt to use AD to compute the Hessian matrix in FWI. \citeA{liao_2011_Accurate} develop a 2D acoustic FWI workflow with the AD tools. \citeA{cao_2015_Computational} validate the method in borehole FWI with synthetic tests. \citeA{zhu_2021_General} obtain gradients in FWI using AD, show the equivalence of AD with the conventional ASM, and develop a preliminary AD-based seismic inversion framework named ADSeismic. \citeA{wang_2021_Elastic} and \citeA{richardson_2023_Deepwave} develop AD-based workflows for elastic wave FWI and least-squares reverse time migration. The integration with deep-learning techniques in FWI can also effectively improve its performance and efficiency \cite{mousavi_2023_Applications}. Recently, some studies explore how to use deep neural networks (DNNs) to reparameterize models for mitigating non-convexity and estimating result uncertainties in FWI \cite{he_2021_Reparameterized, zhu_2022_Integrating, sun_2023_Implicit}. Despite of many endeavors in applying AD to FWI for over a decade, there still lacks a versatile AD-based framework that supports not only wavefield forward simulations in different media, but also corresponding waveform inversion with sophisticated techniques. Moreover, a user-friendly workbench that can facilitate efficient validations of new waveform inversion methods \cite{taufik_2024_Learned, xie_2024_Stochastic} and enable a rapid comparison of different objective functions, optimization algorithms, and regularization techniques is still not readily available to the scientific community. 

    In this work, we present ADFWI, an open-source, AD-based flexible framework that incorporates a variety of state-of-the-art techniques in deep learning and inverse problems for FWI. ADFWI contains not only efficient forward operators for waveform modeling in various types of media, but also flexible inversion strategies with different objective functions, optimization algorithms, and regularization techniques. Also, deep neural networks are integrated and can be used to provide learned constraints in inversion through model reparameterization. Dropout of a neural network can also be used to efficiently assess model uncertainty. 
    
    The paper is structured as follows: we first introduce the theoretical foundation and workflow of AD-based FWI briefly, followed by an overview of the various incorporated objective functions, optimization algorithms, and regularization techniques. Then we show how neural networks are integrated into the proposed framework to impose regularization. Also, several numerical examples for different types of wave equations and subsurface models showcase the robustness and versatility of the framework. Finally, we discuss some issues in AD-based FWI and provide viable solutions.

%
%
\section{The AD-based framework for Full Waveform Inversion}
    \subsection{Formulations for waveform modeling}
        Simulations of seismic waveforms by solving wave equations are key to FWI \cite{fichtner_2010_Full}. The governing equations for the elastic wave propagation are \cite{virieux_1986_PSV, levander_1988_Fourth}:
        \begin{equation}\label{eqn-1}
            \begin{aligned}
                \rho \frac{\partial v_i}{\partial t} &= \frac{\partial \tau_{ij}}{\partial x_j} + f_i
            \end{aligned},
        \end{equation}
        \begin{equation}\label{eqn-2}
            \begin{aligned}
                \tau_{ij} &= C_{ijkl}\epsilon_{kl}
            \end{aligned},
        \end{equation}
        where $\rho$ denotes the density of the medium, $v_i$ denotes the particle velocity, $\tau_{ij}$ and $\epsilon_{kl}$ denote the stress and strain tensors, respectively, $f_i$ represents the external body force in the $i$-th direction, and $C_{ijkl}$ denotes the tensor of elastic moduli, with $i, j, k, l \in \{ x, y, z\}$. Due to the symmetry in the stress and strain components, the fourth-order elastic tensor can be reduced to a second-order tensor using the Voigt notation $C_{ij} (i,j=1,2,...,6)$ \cite{chapman_2004_Fundamentals}. For the vertically transverse isotropy (VTI) or horizontal transverse isotropic (HTI) medium, only 5 independent parameters are needed, which are $C_{11}$, $C_{13}$, $C_{33}$, $C_{44}$, $C_{66}$. The Thomsen's parameters ($\epsilon, \gamma, \delta$, $\alpha_0, \beta_0$) are often chosen to characterize weak anisotropic medium in these cases, where $\alpha_0$ and $\beta_0$ are horizontal P- and S-wave velocities, respectively for the VTI case, or vertical P- and S-wave velocities, respectively for the HTI case, and $\epsilon$, $\gamma$, and $\delta$ determine how the P- and S-wave velocities change with propagation directions \cite{thomsen_1986_Weak, warner_2013_Anisotropic}. In this study, FWI for the VTI- or HTI-anisotropic medium is parameterized with $\alpha_0$, $\beta_0$, $\rho$, $\epsilon$, $\gamma$ and $\delta$. For the isotropic case (ISO), the elastic moduli can be represented by $Lam\acute{e}$' s parameters $\lambda$ and $\mu$, with $C_{11}=C_{33}=\lambda + 2\mu$, $C_{44}=C_{66}=\mu$, $C_{13}=\lambda$. The P- and S-wave velocities that are related to $Lam\acute{e}$'s parameters are inverted for in the isotropic FWI case. While the isotropic or TI elastic wave equations can characterize wave propagation in the real Earth more accurately, the acoustic approximation is often opted for in practice due to its efficiency and more tractable inversion process, especially when inverting seismic data acquired by hydrophones. The governing equations for acoustic wave propagation can be simplified as \cite{alford_1974_ACCURACY, schuster_2017_Seismic}:
        \begin{equation}\label{eqn-3}
            \frac{\partial v_i}{\partial t} = \frac{1}{\rho} \frac{\partial p}{\partial x_i},
        \end{equation}
        \begin{equation}\label{eqn-4}
            \frac{\partial p}{\partial t} = \kappa \nabla \cdot \textbf{v} + f, 
        \end{equation}
        where $p$ denotes the pressure, $f$ is the source term, $\textbf{v}$ represents the particle velocity, and $\kappa$ is the bulk modulus. In acoustic FWI, the inverted parameters of interest are the P-wave velocity $c$, i.e., $c=\sqrt{\kappa /\rho}$, and the density.

        In this study, we use the staggered-grid finite-difference scheme to solve both the ISO acoustic and ISO/VTI elastic wave equations in 2-D with perfectly matched layer boundary conditions \cite{berenger_1994_Perfectly, komatitsch_2003_Perfectly}. The acoustic wave propagator has 4th-order accuracy in space and 2nd-order accuracy in time, and the elastic wave equations are solved with selectable 4th-, 6th-, or 8th-order of accuracy in space, and 2nd-order accuracy in time. Detailed formulas and implementations for those governing equations can be found in \citeA{schuster_2017_Seismic} and \citeA{li_2021_FDwave3D}. 

    \subsection{Automatic Differentiation in FWI: Gradient Calculation and Implementation}
        A general form for the objective function \(\mathcal{J}\) used in FWI can be defined
        \begin{equation}\label{eqn-5}
            \mathcal{J}(\mathbf{m}) = \chi(d_{obs}(s, r, t) , d_{cal}(\mathbf{m}; s, r, t)),
        \end{equation}
        where $\chi$ measures discrepancies between the observed data $d_{obs}(s,r,t)$ and synthetic data $d_{cal}(\textbf{m}; s,r,t)$ for all shots ($s$) and receivers ($r$), and $\textbf{m}$ denotes the model parameters. The misfit can be measured by different norms, such as the L2-norm \cite{lailly_1983_Seismic}, the L1-norm \cite{brossier_2010_Which, guitton_2003_Robust}, the envelope \cite{bozdag_2011_Misfit, wu_2014_seismic}, etc. 
        
        The optimization in FWI can be generally expressed as:
        \begin{equation}\label{eqn-6}
            \textbf{m}^* = \underset{\textbf{m}}{\arg\min}\;(\mathcal{J}(\textbf{m}) + \alpha \mathcal{R}(\textbf{m})),
        \end{equation}
        where $\textbf{m}^*$ indicates the final inverted subsurface model(s), $\mathcal{R}(m)$ represents the regularization term, and $\alpha$ is the corresponding weight. Traditional FWI uses ASM to compute the gradient $\partial J/\partial\textbf{m}$ and updates \textbf{m} with gradient-based optimization methods. However, for governing equations that are not self-adjoint, it can be quite involved to derive the associated adjoint equations and then implement their discretized form numerically. In contrast, AD leverages the chain rule in calculus to compute error-free gradients automatically once the forward modeling process is established \cite{rumelhart_1986_Learning}. By replacing the representation of variables to include derivatives and modifying the semantics of operations to propagate derivatives using the chain rule, AD restructures the forward process into a computation graph \cite{baydin_2015_Automatic}. In the context of FWI, the variables include model parameters ($\textbf{m}$) and intermediate parameters (e.g. the wavefield at each time step), and the operations include wavefield modeling governed by the wave equations and other calculations such as the objective function or regularization. The implicit calculation (i.e., back-propagation) of the gradients of the objective function ($\mathcal{J}$) with respect to each parameter is achieved by tracking the forward calculation process of the computational graph \cite{richardson_2018_Seismic, zhu_2021_General}. In other words, AD eliminates the need to manually derive the adjoint state equations, and once the forward-modeling propagators are constructed with a computational graph, calculations of the gradients are then handled automatically by AD. 
        
        Unlike symbolic differentiation which can be cumbersome for complex functions \cite{grabmeier_2003_Computer}, or numerical differentiation which approximates gradients \cite{frolkovic_1990_Numerical}, AD provides a satisfying measure to compute derivatives rigorously and efficiently for arbitrary differential functions. Previous studies validate the accuracy of gradients by AD through theoretical analyses \cite{zhu_2021_General} and synthetic tests \cite{cao_2015_Computational, richardson_2023_Deepwave}. In addition, unlike in ASM where the adjoint sources need to be derived and implemented depending on the choice of the objective function, the AD framework does not need explicit adjoint sources. With all the advantages for AD-based FWI, however, construction of complex computational graphs was not readily available in traditional AD tools. With rapid progress in modern deep learning libraries such as PyTorch \cite{paszke_2017_Automatic}, construction and management of computational graphs can be automatically processed, and thus the complexity of building such graphs for intricate inverse problems have been significantly reduced. A simple validation of the accuracy of the AD-based gradients is shown in Supporting Figure S1.

    \subsection{ADFWI: Automatic Differentiation-Based Full Waveform Inversion}
        Based on PyTorch, we develop ADFWI which is an AD-based, flexible framework dedicated to efficient implementations, evaluations, and comparisons of different approaches in full waveform inversion. ADFWI is capable of modeling wave propagation in isotropic acoustic (ISO-acoustic), isotropic elastic (ISO-elastic), and TI elastic (VTI- and HTI-elastic) media, as well as corresponding waveform-based inversion with various objective functions, optimization algorithms and regularization techniques (Figure 1). In this framework, strategies for addressing challenges such as cycle skipping, local minima and non-uniqueness can be readily explored. In addition, ADFWI leverages DNN and dropout in deep learning to introduce learned regularization and uncertainty estimation, which are not available in traditional FWI.

        \clearpage
        
        \begin{figure}[!ht]
            \centering
            \includegraphics[width=0.95\textwidth]{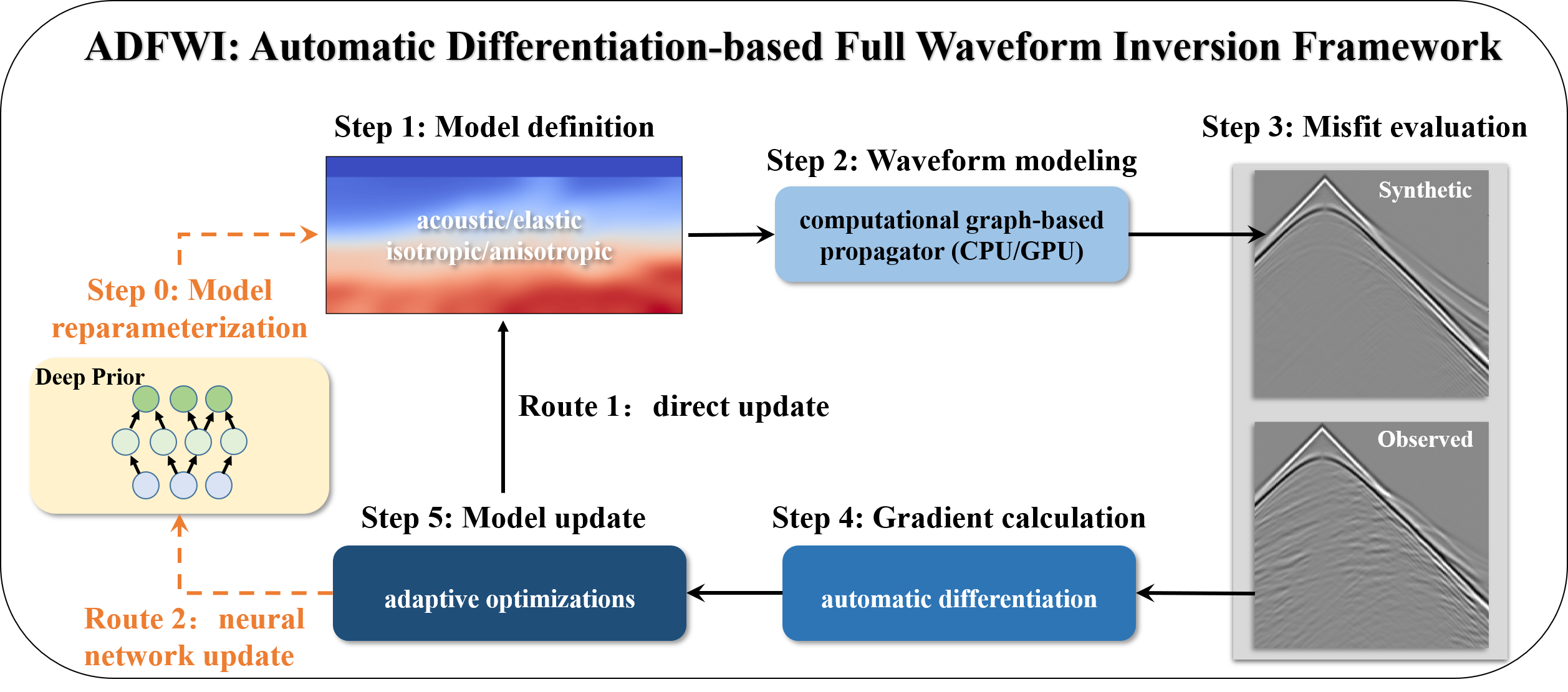}
            \caption{Workflow of ADFWI, which includes five essential steps with an optional neural network-based reparameterization/regularization step. Step 1: define the isotropic or anisotropic acoustic/elastic Earth model and the data acquisition system. Step 2: perform forward wavefield modeling using propagators implemented in Pytorch on CPU or GPU. Step 3: measure the discrepancy between observed and synthetic waveforms based on the objective function of choice. Step 4: compute gradients of the objective function with respect to model parameters via automatic differentiation. Step 5: update model parameters using the computed gradients with optimization schemes of choice. Optional step 0: re-parameterize the Earth model using a deep neural network (DNN). When the model is parameterized using a DNN, the neural network parameters are updated during the optimization (Route 2); otherwise, the velocity model parameters are updated directly (Route 1).}
            \label{Figure1}
        \end{figure}
        
        \vspace{-12pt}
        
        \subsubsection{Objective Functions}
            Objective functions are crucial for properly quantifying the discrepancy between synthetic and observed seismic data, and the choice can significantly impact misfit convergence, computational efficiency, and mitigation of cycle skipping in FWI. Traditional metrics such as the L2-norm are commonly used, which however, is sensitive to initial models, prone to be affected by non-Gaussian noise, and trapped in local minima \cite{tarantola1984linearized}. The envelope-based objective function provides improved convexity and the ability to capture low-wavenumber structures in the early iterations, but has limitations in resolving structural details in later iterations \cite{wu_2014_seismic}. Many novel objective functions have been introduced for FWI recently, such as dynamic time warping (DTW) and the Wasserstein distance, which were originally developed for time series analysis and image processing. These new metrics for measuring data misfit can offer advantages in addressing inherent nonlinearity in FWI and have shown promising results \cite{ma_2013_wave, engquist_2016_Optimal}. In ADFWI, we integrate the following objective functions, which are categorized into three groups.
            \begin{enumerate}
                \item \textbf{Waveform based:} these objective functions directly measure the difference between the observed and synthetic waveforms, including:
                \begin{itemize}
                    \item \textbf{L2-norm (L2):} commonly used due to its simplicity and effectiveness, but is sensitive to amplitude errors and prone to cycle-skipping \cite{lailly_1983_Seismic, virieux_2009_Overview}.
                    \item \textbf{L1-norm (L1):} more robust against outliers, with improved stability for noisy data \cite{brossier_2010_Which, guitton_2003_Robust}.
                    \item \textbf{T-Distribution (StudentT):} balanced data sensitivity and robustness, especially effective in handling data with outliers \cite{aravkin_2011_Robust, guo_2023_Leastsquares}.
                \end{itemize}
                
                \item \textbf{Waveform-attributes based:} these objective functions leverage various attributes of seismic waveforms to enhance robustness and convergence of the inversion process, albeit at the cost of potentially reduced resolution in the inversion results, including: 
                \begin{itemize}
                    \item \textbf{Envelope:} focused on amplitude variations instead of phases, with boosted low-frequency contents to improve resistance to noise, effective for data with low signal-to-noise ratio \cite{bozdag_2011_Misfit, wu_2014_seismic}.
                    \item \textbf{Global correlation (GC):} aligned phase information between observed and synthetic data, and improved convergence by mitigating amplitude-related discrepancies \cite{choi_2012_Application}.
                \end{itemize}
                
                \item \textbf{Data-alignment based:} these objective functions compare observed and synthetic data using more advanced alignment methods, including both waveform features and statistical distributions. Traditional forms of these functions involve complex computations, and the corresponding adjoint sources in the ASM framework can be difficult to implement. Additionally, the non-differentiable operations in the objective functions can break gradient back-propagation by disrupting the chain rule in AD. Recent advancements in deep learning have adopted modifications to allow efficient gradient back-propagation for these objective functions, which include:
                \begin{itemize}
                    \item \textbf{Differentiable dynamic time warping (Soft-DTW):} differentiable DTW for measuring the similarity between two time-dependent sequences, effective in aligning time-shifted waveforms \cite{ma_2013_wave, cuturi_2017_SoftDTW, chen_2022_Cycleskipping}.
                    \item \textbf{Wasserstein distance with Sinkhorn divergence (Wasserstein-Sinkhorn):} quantifying the minimal transport cost between observed and synthetic data distributions with Sinkhorn regularization, providing a robust measurement for distributional alignment trace by trace \cite{engquist_2016_Optimal, metivier_2016_Measuring, yang_2018_Application, chizat_2020_Faster}.
                \end{itemize}
            \end{enumerate}
            
            Each group of objective functions is specialized in capturing certain attributes of seismic waveform data. By leveraging AD, different objective functions can be implemented with trivial efforts, since only the value of the objective function is calculated, but the corresponding adjoint source needs not to be derived and implemented. Therefore, not only existing objective functions can be easily compared, but also newly designed ones can be readily implemented in ADFWI. Further details for these objective functions are provided in Supporting Text S1.
            
        \subsubsection{Adaptive Gradient Optimization Methods}
            Once gradients are obtained by AD, the model parameters can be updated using gradient optimization algorithms. The iterative optimization methods in conventional FWI often include the nonlinear conjugate-gradient method (NLCG) and quasi-Newton methods such as the limited-memory Broyden-Fletcher-Goldfarb-Shanno algorithm (\textit{l}-BFGS) \cite{fletcher_1964_Function, nocedal_1980_Updating}. Recently, another type of optimization method, the adaptive gradient optimization (AGO) gained popularity in both the FWI and deep learning communities due to superior computational efficiency \cite{bernal-romero_2020_Accelerating}. As conventional optimization such as NLCG or \textit{l}-BFGS methods have been well investigated in FWI \cite{modrak_2016_seismic},
            in this study we benchmark six different AGO methods for the AD-based FWI, which are listed below.
            \begin{itemize}
                \item \textbf{Root mean squares propagation (RMSProp)}: adjusts the update rate for each parameter based on the recent average of the squared gradients, more stable when updating with noisy gradients \cite{graves_2014_Generating}.
                
                \item \textbf{Adaptive gradient algorithm (Adagrad)}: adjusts the update rate for each parameter individually based on the historical gradients, improving update efficiency in scenarios where some parameters are updated less frequently \cite{duchi_2011_Adaptive}.
                
                \item \textbf{Adaptive moment estimation (Adam)}: combines the advantages of both RMSProp and momentum for improved convergence by maintaining a moving average for both the gradient and its squares \cite{kingma_2017_Adam}.
                
                \item \textbf{Adam with weight decay (AdamW)}: modifies Adam by decoupling weight decay from the optimization steps, providing better regularization for model parameters \cite{loshchilov_2019_Decoupled}.
                
                \item \textbf{Nesterov-accelerated adaptive moment estimation (NAdam)}: integrates Nesterov's accelerated gradient into Adam, providing faster convergence by incorporating a lookahead mechanism in the update rules \cite{dozat_2016_Incorporating}.
                
                \item \textbf{Rectified Adam (RAdam)}: introduces a rectification mechanism to Adam, stabilizing model updates by correcting the variance of the adaptive learning rate in early iterations \cite{liu_2019_Variance}.
            \end{itemize}
            
            In addition, the \textit{l}-BFGS algorithm is also integrated in ADFWI. Further details for the incorporated AGO methods can be found in Supporting Text S2.
        
        \subsubsection{Regularization Techniques}
            Regularization is crucial for successful full waveform inversion which is inherently ill-posed. Incorporation of appropriate constraints or priors can expedite convergence and guide model updates towards geologically feasible solutions. In ADFWI, we integrate two commonly used regularization techniques, the total variation (TV) and Tikhonov regularizations \cite{rudin_1992_Nonlinear, engl_2000_Regularization, vogel_2002_Computational}. The TV regularization is effective for preserving sharp interfaces and discontinuities, which are critical for capturing lithological boundaries \cite{modrak_2016_seismic}. In comparison, the Tikhonov regularization imposes smoothness and structural continuity in the model, which enhances inversion stability and avoid overfitting noisy data. While we only showcase the performances of two traditional regularization techniques integrated in ADFWI, it should be emphasized that the proposed framework allows for convenient implementations of additional advanced regularization techniques, such as the hybrid \cite{aghamiry_2020_Compound} and deep learning-based regularization \cite{sun_2023_FullWaveform}.
    
        \subsubsection{Reparameterization with Deep Neural Network}
            Originally developed for image processing tasks, deep image prior (DIP) has been successfully adapted for FWI in reparameterization of subsurface models using the inherent structure and parameters of DNN \cite{he_2021_Reparameterized, zhu_2022_Integrating, sun_2023_FullWaveform, wang_2023_Prior}. The DNN-based reparameterization can readily impose implicit model regularization and embed physical constraints, with no requirement for external training data \cite{ulyanov_2018_Deep}. As illustrated in Figure 1, a model can be alternatively reparameterized by a neural network in ADFWI. The neural network generates a velocity model (Step 0), followed by waveform modeling based on this model and computation of the misfit between synthetic and observed waveforms. Unlike previous FWI workflows integrated with DIP \cite{he_2021_Reparameterized} which separate neural network training from the model optimization, the new framework in ADFWI uses AD to directly compute gradients of the objective function with respect to the parameters of the neural network (Route 2 in Figure 1). With reparameterization, the velocity model is expressed as:
            \begin{equation}\label{eqn-7}
                \mathbf{m} = \mathcal{N}(z_l, \omega),
            \end{equation}
            where $\mathcal{N}$ represents a neural network with latent variables $z_l$ and weights $\omega$. Accordingly, the objective function (Eq. 6) can be reformulated as:
            \begin{equation}\label{eqn-8}
                \omega^* = \underset{\omega}{\arg \min}\;(\mathcal J(\mathcal{N}(z_l, \omega); m_0) + \alpha \mathcal{R}(\mathbf{m})),
            \end{equation}
            where $m_0$ is the initial model. In this context, the DNN serves as an implicit prior, and its architecture (e.g., CNNs, U-Nets) can encode structural information of the velocity model (Figure 2). To address the challenge in generating reasonable velocity models from a randomly initialized neural network \cite{sun_2023_Implicit}, a pre-training strategy is implemented prior to conducting FWI. This approach allows the network $\mathcal{N}$ to first learn an end-to-end mapping from prompt pre-training between random input vectors and an initial velocity model derived from, e.g., travel-time inversion \cite{he_2021_Reparameterized, sun_2023_Implicit}. After pre-training, the gradients from AD iteratively update the network $\mathcal{N}$, which first improves the low-wavenumber features in the velocity model, followed by refinement of the higher-wavenumber structures \cite{shi_2021_Measuring, ulyanov_2018_Deep}. Model smoothing can be controlled by the number of iterations, a technique commonly referred to as early stopping in deep learning \cite{ulyanov_2018_Deep}.
            
            It is important to note that ADFWI incorporates a variety of network architectures for model reparametrization, including multilayer perceptrons \cite{rumelhart_1986_Learning}, multilayer CNNs \cite{lecun_1998_Gradientbased}, U-shaped networks \cite{ronneberger_2015_Unet}, and residual networks \cite{he_2016_Deep}. Additionally, ADFWI can incorporate other network architectures, such as visual transformers \cite{dosovitskiy_2020_Image} and diffusion models \cite{ho_2020_Denoising}. In the following, we show an example where DNN is used to reparameterize the inverted model with sophisticated regularization.
            
            \begin{figure}[!ht]
                \centering
                \includegraphics[width=1.0\textwidth]{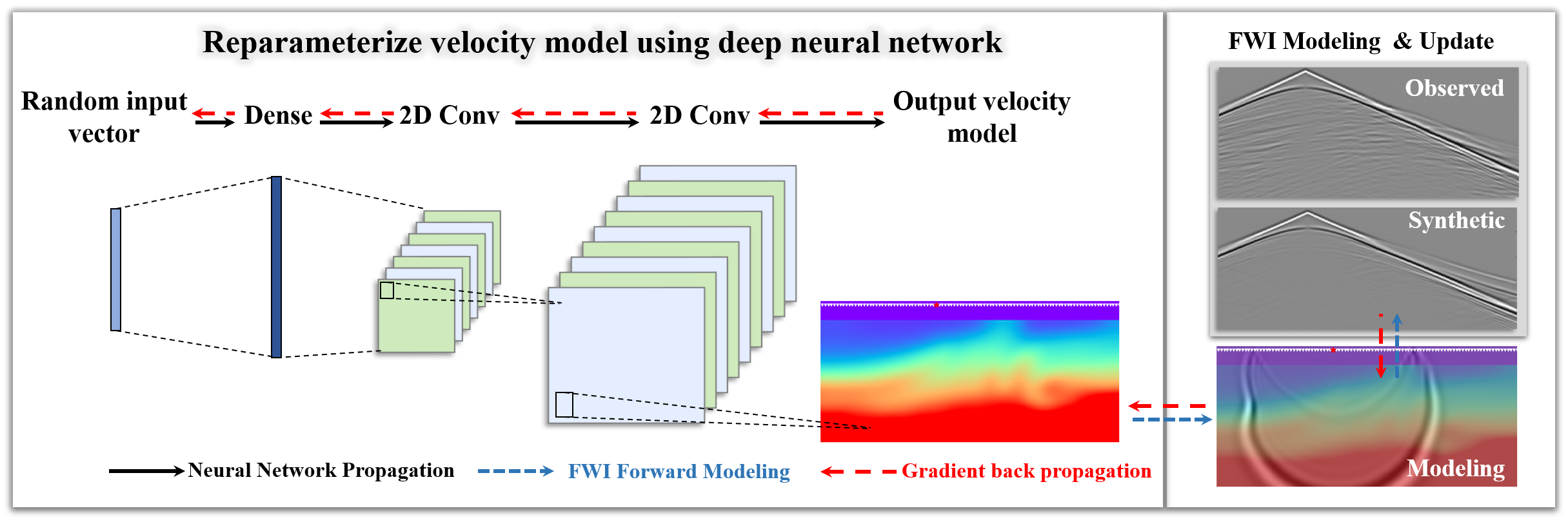}
                \caption{FWI workflow with model reparameterization using a deep neural network. The process begins by generating a velocity model from a random vector using a generative network, which typically is a multilayer convolutional neural network. The velocity model is then used in the forward propagator for waveform modeling, and the gradients from AD are then back-propagated to update the neural network.}
                \label{Figure2}
            \end{figure}

%
%
\section{Results}

    \subsection{Evaluation of ADFWI with various types of wave equations}

        We first conduct a series of synthetic tests to evaluate the performance of ADFWI in ISO-acoustic, ISO-elastic, and VTI-elastic media. In these tests, we use the Ricker wavelet as the source signature, the L2-norm-based objective function to measure waveform discrepancy, and the Adam optimizer for iterative updates. Other key parameters for these tests are listed in Table 1.
                       
        \subsubsection{ISO-acoustic FWI with the Marmousi2 and Overthrust Models}
        
            We first evaluate ADFWI using the ISO-acoustic Marmousi2 \cite{martin_2006_Marmousi2} and Overthrust models \cite{aminsadeh_1996_3D}. The Marmousi2 model is resized to 200 $\times$ 88 grids in the $x$ and $z$ directions, with a uniform grid interval of 40 m. The true P-wave velocity model is shown in Figure 3a. A total of 40 sources and 200 receivers are placed along the surface. The other key parameters for the synthetic tests can be found in Table 1. The inversion starts with a smoothed version of the true model by using a Gaussian smoothing window (Figure 3b), and the velocity in the water layer remains fixed during the inversion. The density model is derived from the velocity model with the empirical formulas \cite{gardner_1974_FORMATION}. Figure 3c shows the inverted model after 300 iterations, which contains nicely recovered heterogeneities. Similarly, the test for the Overthrust model is shown in Figures 3d-f. In particular, the alternating interbedded high and low velocities are also well recovered (Figure 3f). Note that only $v_p$ is inverted here to facilitate comparisons with subsequent tests. The example for simultaneous inversion of both the $v_p$ and $\rho$ models is shown in Supporting Figure S2.

            \begin{table}[!ht]
                \caption{Key parameters of the tests for ADFWI in various types of media. In all tests a Ricker wavelet is used, though the dominant frequency varies.}
                \centering
                \resizebox{\textwidth}{!}{%
                \begin{tabular}{l|cccc}
                    \hline
                    &
                    \textbf{\begin{tabular}[c]{@{}c@{}}ISO-acoustic\\ Marmousi2\end{tabular}} &
                    \textbf{\begin{tabular}[c]{@{}c@{}}ISO-acoustic\\ Overthrust\end{tabular}} &
                    \textbf{\begin{tabular}[c]{@{}c@{}}ISO-elastic\\ Marmousi2\end{tabular}} &
                    \textbf{\begin{tabular}[c]{@{}c@{}}VTI-elastic anomaly\end{tabular}}\\ \hline
                    \textbf{Grid spacing} & 40\,m & 50\,m & 45\,m & 10\,m \\
                    \textbf{Time interval} & 0.003\,s & 0.003\,s & 0.003\,s & 0.001\,s \\
                    \textbf{Number of grids} & 200 $\times$ 88 & 200 $\times$ 100 & 200 $\times$ 80 & 180 $\times$ 80 \\
                    \textbf{Time steps} & 1600 & 1600 & 2500 & 1000 \\
                    \textbf{Number of receivers} & 200 & 200 & 200 & 180 \\
                    \textbf{Number of sources} & 40 & 40 & 40 & 36 \\
                    \textbf{\begin{tabular}[c]{@{}l@{}}Source dominant freq.\end{tabular}} & 5\,\text{Hz} & 5\,\text{Hz} & 3\,\text{Hz} & 30\,\text{Hz} \\
                    \textbf{Smooth window size} & 240\,\text{m} $\times$ 240\,\text{m} & 500\,\text{m} $\times$ 500\,\text{m} & 180\,\text{m} $\times$ 180\,\text{m} & — \\
                    \textbf{Inverted parameters} & $v_p$ ($\rho$) & $v_p$ & $v_p$, $v_s$ & $\epsilon$, $\delta$ \\
                    \hline
                \end{tabular}%
                }
            \end{table}
            
            \vspace{-6pt}
            
            \begin{figure}[!ht]
                \centering
                \includegraphics[width=1.0\textwidth]{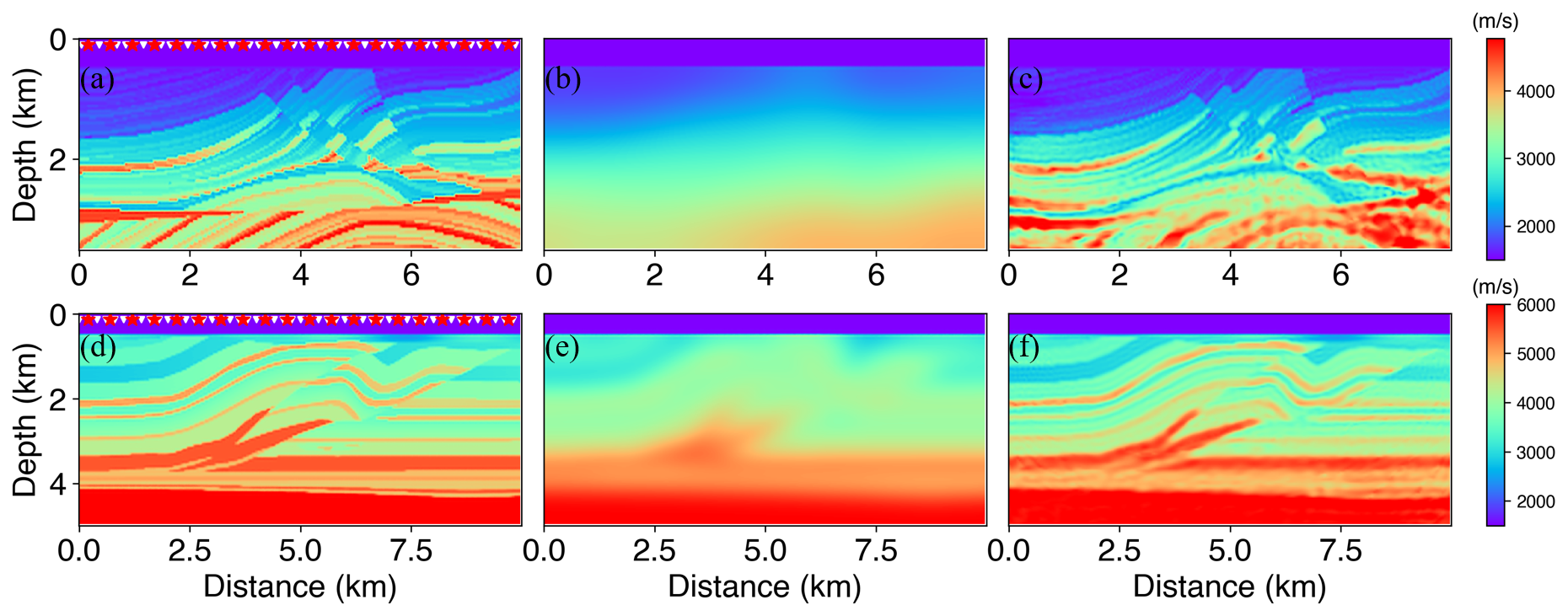}
               \caption{FWI tests using the ISO-acoustic Marmousi2 and Overthrust models. (a) The true Marmousi2 model, where the red stars indicate source locations and the white triangles indicate the receiver positions; (b) the initial model, obtained by applying a 240 $\times$ 240 m Gaussian smoothing filter to the true model; (c) inverted Marmousi2 model after 300 iterations; (d)-(f) are similar to (a)-(c), but for the Overthrust model. The initial model is generated by applying a 500 $\times$ 500 m Gaussian smoothing filter to the true model.}
                \label{Figure3}
            \end{figure}
            
            \clearpage
            
            \subsubsection{ISO-elastic FWI with the Marmousi2 Model}

                We further evaluate the performance of ADFWI in ISO-elastic media using the elastic version of the Marmousi2 model \cite{martin_2006_Marmousi2}. In this case, we invert for both the P- and S-wave velocities simultaneously. The model parameters and the configurations for the sources and receivers are summarized in Table 1. The true P-wave velocity ($v_p$) and S-wave velocity ($v_s$) are shown in Figures 4a and 4b, and the corresponding initial models obtained by applying a 180 $\times$ 180 m Gaussian smoothing window to the true models are shown in Figures 4c and 4d. Given the complicated trade-offs among multiple parameters and the low sensitivity to density in elastic FWI \cite{tarantola_1986_Strategy, virieux_2009_Overview}, the $\rho$ model is assumed to be a constant of 2.45 $g/cm^3$ in this case, though it can be inverted for in ADFWI nevertheless. Figures 4e and 4f show that ADFWI effectively resolves the elastic properties of the model. While the reconstructed $v_p$ and $v_s$ models are more accurate in the shallow regions, the inversion results at greater depths are less satisfactory, reflecting the inherent challenges of elastic FWI \cite{tarantola_1986_Strategy, Brossier_2009_Seismic}. Techniques such as gradient preconditioning can be used to improve the accuracy and resolution of structures at greater depths \cite{virieux_2009_Overview}. Also, while the Adam optimizer has proven to be effective for acoustic FWI \cite{richardson_2018_Seismic, bernal-romero_2020_Accelerating, zhu_2022_Integrating}, careful tuning of the optimizer, together with gradient preconditioning, is still required to ensure proper model updates in elastic FWI \cite{wang_2021_Elastic}.

                \begin{figure}[!ht]
                    \centering
                    \includegraphics[width=0.9\textwidth]{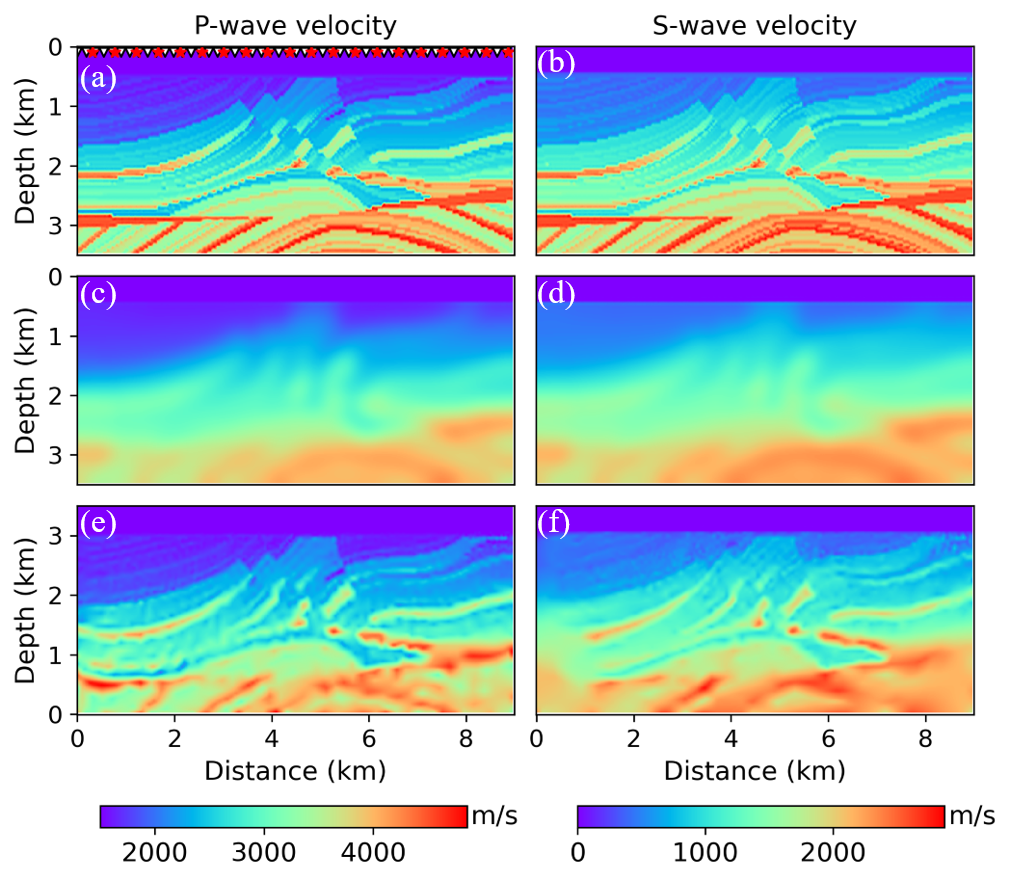}
                    \caption{Test for ADFWI using the ISO-elastic Marmousi2 model. (a) The true P-wave velocity ($v_p$), and (b) true S-wave velocity ($v_s$) of the elastic Marmousi2 model. The red stars indicate source locations, and the white inverted triangles indicate the receiver positions; (c)-(d) the initial models for $v_p$ and $v_s$, respectively, obtained by applying a 180 $\times$ 180 m Gaussian smoothing window to the true models; (e) the inverted $v_p$, and (f) $v_s$ model after 300 iterations.}
                    \label{Figure4}
                \end{figure}
                
            \subsubsection{VTI-elastic FWI with Anomaly Model}

                In this section, we benchmark ADFWI in a VTI model containing multiple anomaly inclusions with diverse shapes and degrees of anisotropy (Figure 5). In this case, $\epsilon$ is 0.1 for the background, while $\epsilon$ in the anomaly inclusions range from 0.15 to 0.3 (Figures 5a and 5c). Similarly, $\delta$ is 0.05 for the background, and in the anomaly inclusions $\delta$ ranges from 0.1 to 0.25 (Figures 5b and 5d). For simplicity and clarity, the test is specifically designed to validate the reconstruction of $\epsilon$ and $\delta$, and thus correct $v_p$ and $v_s$ which are 3000 m/s and 1500 m/s, respectively, for the entire model are used in the inversion. It should be emphasized that $\epsilon$ and $\delta$ are inverted for simultaneously in this test. As shown in Figures 5e and 5f, ADFWI successfully recovers the $\epsilon$ and $\delta$ models with trivial crosstalk.

                \begin{figure}[!ht]
                    \centering
                    \includegraphics[width=0.9\textwidth]{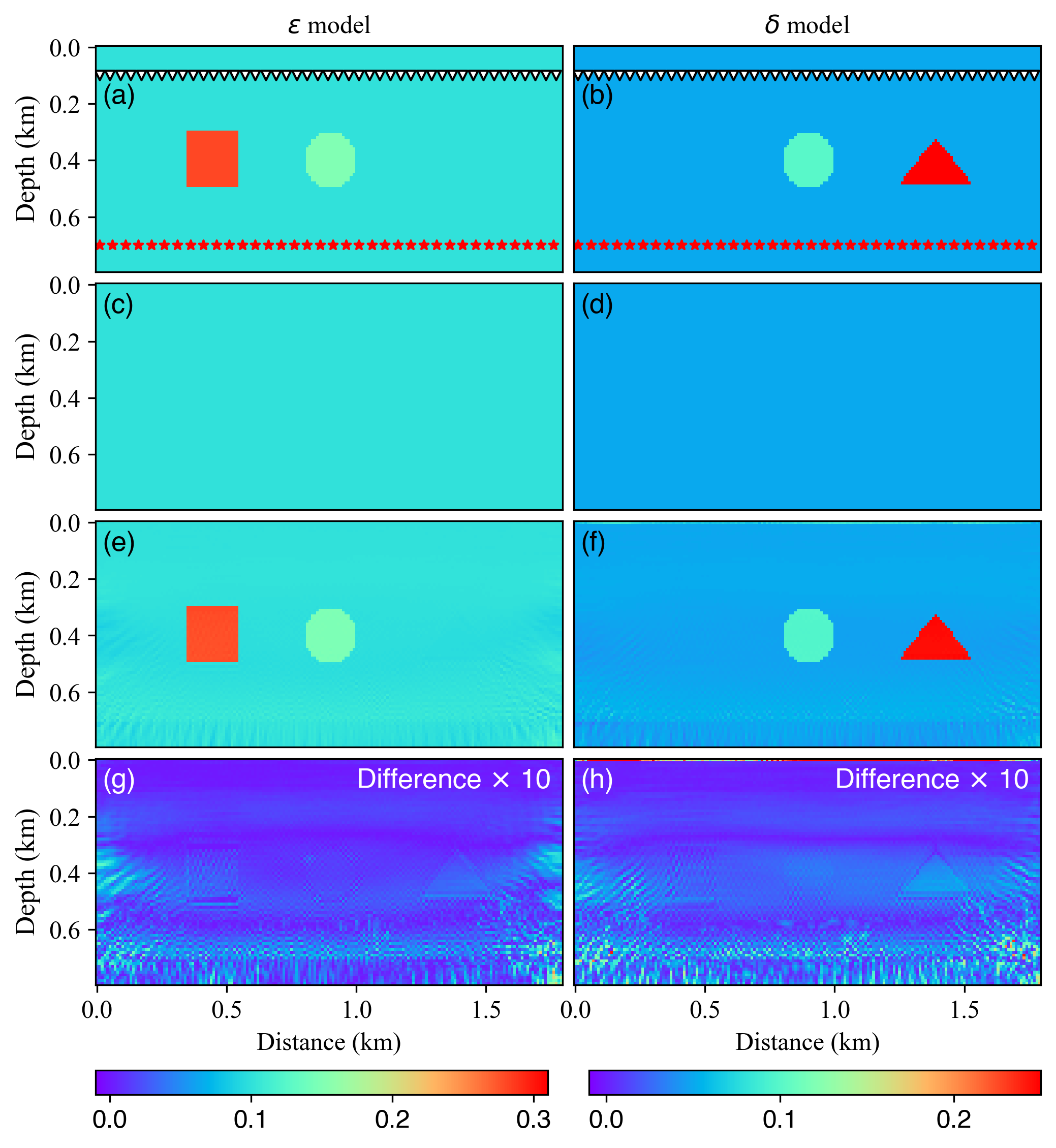}
                    \caption{VTI-elastic FWI tests using a model containing multiple anomaly inclusions with perturbed $\epsilon$ and $\delta$. (a)-(b) The true $\epsilon$ and $\delta$ models, where red stars indicate source locations and white inverted triangles mark receiver positions. The background values are $\epsilon$=0.1 and $\delta$=0.05, with anomaly inclusions having $\epsilon$ ranging from 0.15 to 0.3 and $\delta$ from 0.1 to 0.3. The entire model has $v_p$=3000 m/s and $v_s$=1500 m/s. (c)-(d) Initial models for $\epsilon$ and $\delta$, which are set to constant values of 0.1 and 0.05, respectively. (e)-(f) Inverted $\epsilon$ and $\delta$ models after 500 iterations using ADFWI. (g)-(h) Differences between the true and inverted $\epsilon$ and $\delta$ models, which are amplified by a factor of 10 to emphasize details.}
                    \label{Figure5}
                \end{figure}
    
    \subsection{Evaluation of the Objective Functions and Optimizers in ADFWI}
        
        \subsubsection{Comparison of Various Objective Functions}
        
            Benefiting from rapid development in deep learning, many complex or even non-differentiable objective functions have been efficiently implemented in libraries such as PyTorch \cite{cuturi_2017_SoftDTW, feydy_2019_interpolating}. Other advanced features such as customized objective functions and regularization are also available in the well-developed libraries and can be readily incorporated for FWI studies. Also, in the new framework it is trivial to combine multiple objective functions to take their respective advantages \cite{vyas_2022_FWI}. To intuitively illustrate the varying convexity of different objective functions, we conduct a comparative analysis using a time-shifted Ricker wavelet. Figure S3 shows the normalized misfits for the seven integrated objective functions, which exhibit varying sensitivities to the time shift. The traditional metrics such as the L1-norm, L2-norm, Student-T, and global-correlation exhibit multiple local minima due to cycle skipping. The envelope-based and data-alignment-based objective functions (Soft-DTW and Wasserstein-Sinkhorn) exhibit better convexity in comparison. 
            
            The performances of the objective functions are further compared using the ISO-acoustic Marmousi2 model as used in the previous section. The initial model is derived by applying a larger Gaussian smoothing window (480 $\times$ 480 m) to intentionally generate cycle skipping. Figure 6 shows the inversion results with different objective functions after 300 iterations with the same Adam optimizer. The comparison of the extracted velocity profiles from different inverted models is shown in Figure S4, which shows that inversions with the traditional objective functions are more likely to fall into local minima than those with Soft-DTW and Wasserstein distance. We further use the mean absolute percentage error (MAPE) \cite{hyndman_2006_Another} and the structural similarity index measure (SSIM) \cite{wang_2004_Image} to quantify the discrepancy between the inverted and the true models:
            \begin{equation}\label{eqn-9}
                MAPE(v, \hat{v}) = \frac{1}{m \times n} \sum_{i = 1}^{m} \sum_{j = 1}^{n} \left| \frac{v_{i,j} - \hat{v}_{i,j}}{v_{i,j}} \right| \times 100\%,
            \end{equation}
            \begin{equation}\label{eqn-10}
                SSIM(v, \hat{v}) = \frac{(2 \mu_{v} \mu_{\hat{v}} + c_1)(2 \sigma_{v\hat{v}} + c_2)}{(\mu_{v}^2 + \mu_{\hat{v}}^2 + c_1)(\sigma_{v}^2 + \sigma_{\hat{v}}^2 + c_2)},
            \end{equation}
            where \(m\) and \(n\) are the sizes of the model, \(v_{i,j}\) and \(\hat{v}_{i,j}\) are the true and inverted velocity models, respectively, \(\mu_{v}\) and \(\sigma_{v}\) are the local mean and standard deviation for the true model, \(\mu_{\hat{v}}\) and \(\sigma_{\hat{v}}\) are those for the inverted model, and \(\sigma_{v,\hat{v}}\) is the cross-covariance between the true and inverted models. Figure 7 shows variations of SSIM with iterations for different optimizers. More details for the comparison can be found in Table 2.
            
            \begin{figure}
                \centering
                \includegraphics[width=1.0\textwidth]{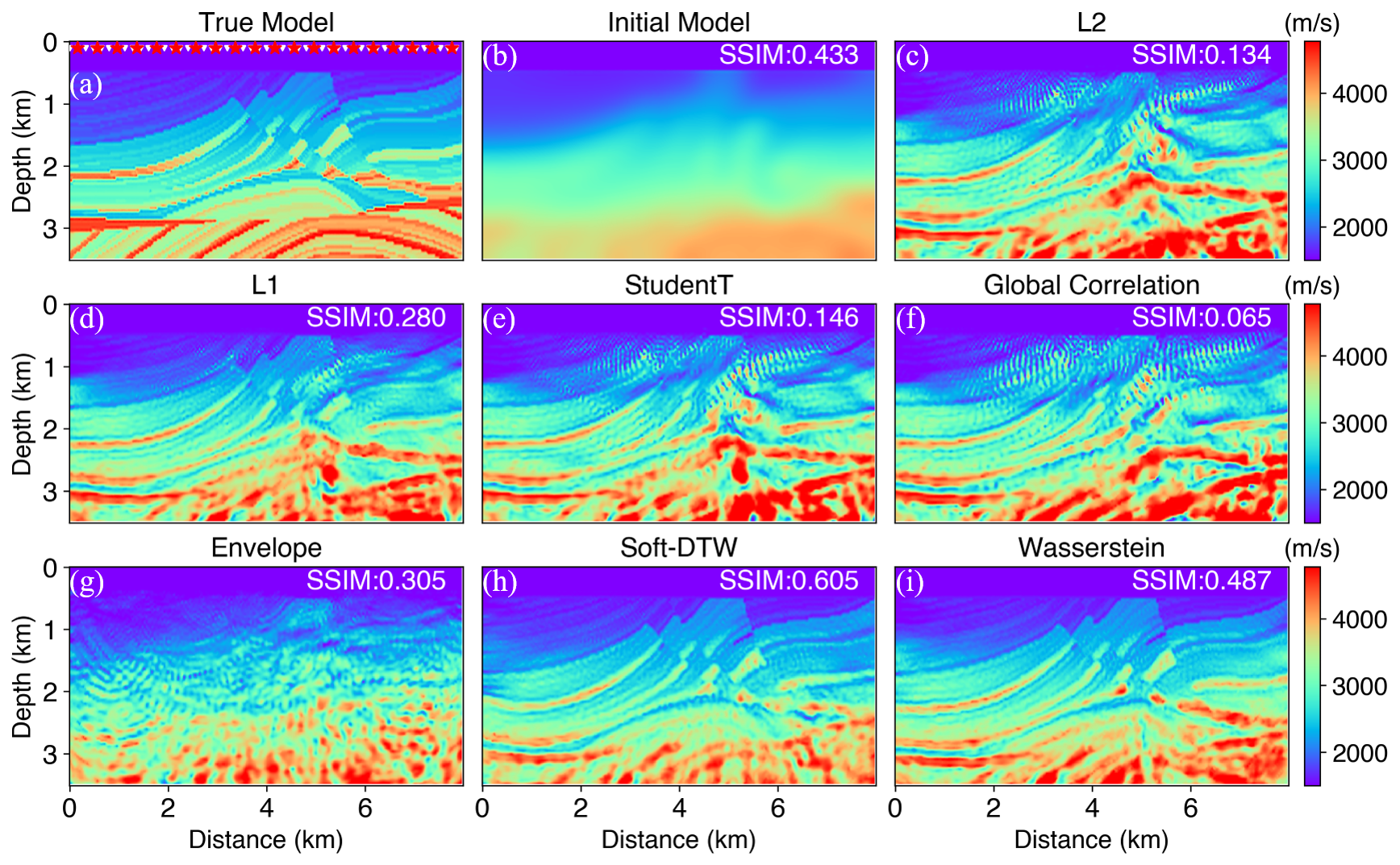}
                \caption{Comparison of the inversion results using various objective functions. (a) The true P-wave velocity model, with the red stars indicating sources and white triangles indicating receivers. (b) The initial P-wave velocity model obtained by applying a 480 $\times$ 480 m Gaussian smoothing window to the true models. Inversion results using the (c) L2-norm, (d) L1-norm, (e) Student's t-distribution, (f) global correlation, (g) envelope, (h) soft-DTW, and (i) Wasserstein distance. Note when evaluating the SSIM which is shown in each subfigure, the identical water layers are excluded.}
                \label{Figure6}
            \end{figure}

            \begin{figure}
                \centering
                \includegraphics[width=0.85\textwidth]{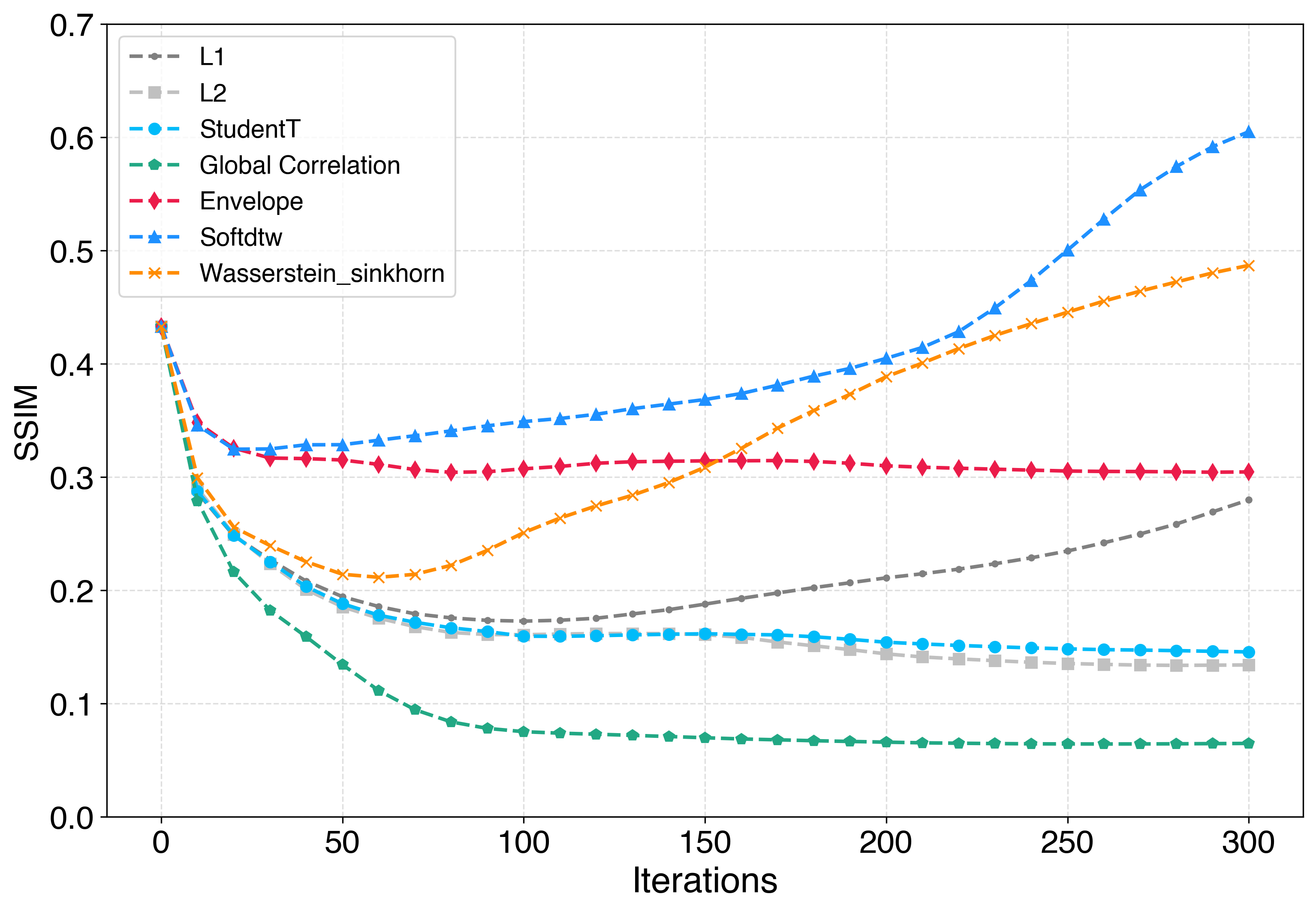}
                \caption{Variations of model similarities (SSIM, higher is better) with iterations for different objective functions.}
                \label{Figure7}
            \end{figure}

            \begin{table}[!ht]
                \caption{Comparison of the MAPE and SSIM for different objective functions. The best results are highlighted in bold. For MAPE, lower values indicate better models, while for SSIM higher values indicate better models.}
                \resizebox{\textwidth}{!}{%
                    \begin{tabular}{c|ccccccc}
                        \hline
                        \diagbox{Metric}{Objective} & L1 & L2 & StudentT & Global corr. & Envelope & Soft-DTW & Wasserstein Dist. \\ \hline
                        MAPE & 12.607 &	15.465 & 15.486 & 17.094 & 11.034 &	\textbf{7.605} & 9.692 \\
                        SSIM & 0.280 & 0.134 & 0.146 & 0.065 & 0.305 & \textbf{0.605} & 0.487 \\ \hline
                        \end{tabular}
                }
            \end{table}

        \clearpage
        
        \subsubsection{Comparison of Various Adaptive optimization methods}

            The choice of the optimization algorithm influence the convergence rate, inversion accuracy, and how nonlinearity and local minima are handled \cite{bernal-romero_2020_Accelerating}. The well-developed optimizers from PyTorch are naturally inherited in ADFWI to facilitate inversion. Here we compare two stochastic gradient descent (SGD) methods and six adaptive optimization (AGO) methods using the ISO-acoustic Marmousi2 model. The model setup is detailed in Section 3.1.1. The L2-norm is used for all inversions in this test, and the hyperparameters for each optimizer are tuned to ensure as fair a comparison as possible. Figure 8 shows the inversion results with various optimizers after 300 iterations, and the detailed metrics are presented in Table 3. Figure S5 further shows the extracted velocity profiles for detailed comparisons. Figures 9a and 9b show the data residual and model similarity (SSIM) with iterations, which suggests the effectiveness of different optimizers varies markedly. While simpler methods like SGD and averaged SGD (ASGD) manage to capture certain structural characteristics, their convergence speed and the final inversion results are much inferior compared to those obtained with the AGO algorithms. Except for Adagrad, all AGO algorithms demonstrate rapid convergence in the data domain and satisfactory improvement in the model domain. In particular, Adam, AdamW, and NAdam achieve the fastest convergence rates in the data domain, which can be attributed to the dynamically adjusted learning rates \cite{dozat_2016_Incorporating, kingma_2017_Adam, loshchilov_2019_Decoupled}. However, it should be emphasized that the effectiveness of optimizers can vary significantly with different datasets and target models. Furthermore, it should be mentioned that ADFWI also incorporates the \textit{l}-BFGS algorithm. However, due to the internal line search in \textit{l}-BFGS, its parameter setting and number of iterations differ from those in other optimizers. Therefore, a separate comparison between \textit{l}-BFGS and AdamW is presented separately in Figure S6 to provide a reasonably unbiased assessment. Overall, the evaluations suggest the optimizers originally designed for machine learning are rather effective, easy to use, and highly suitable for FWI.
            
            \begin{figure}[!ht]
                \centering
                \includegraphics[width=1.0\textwidth]{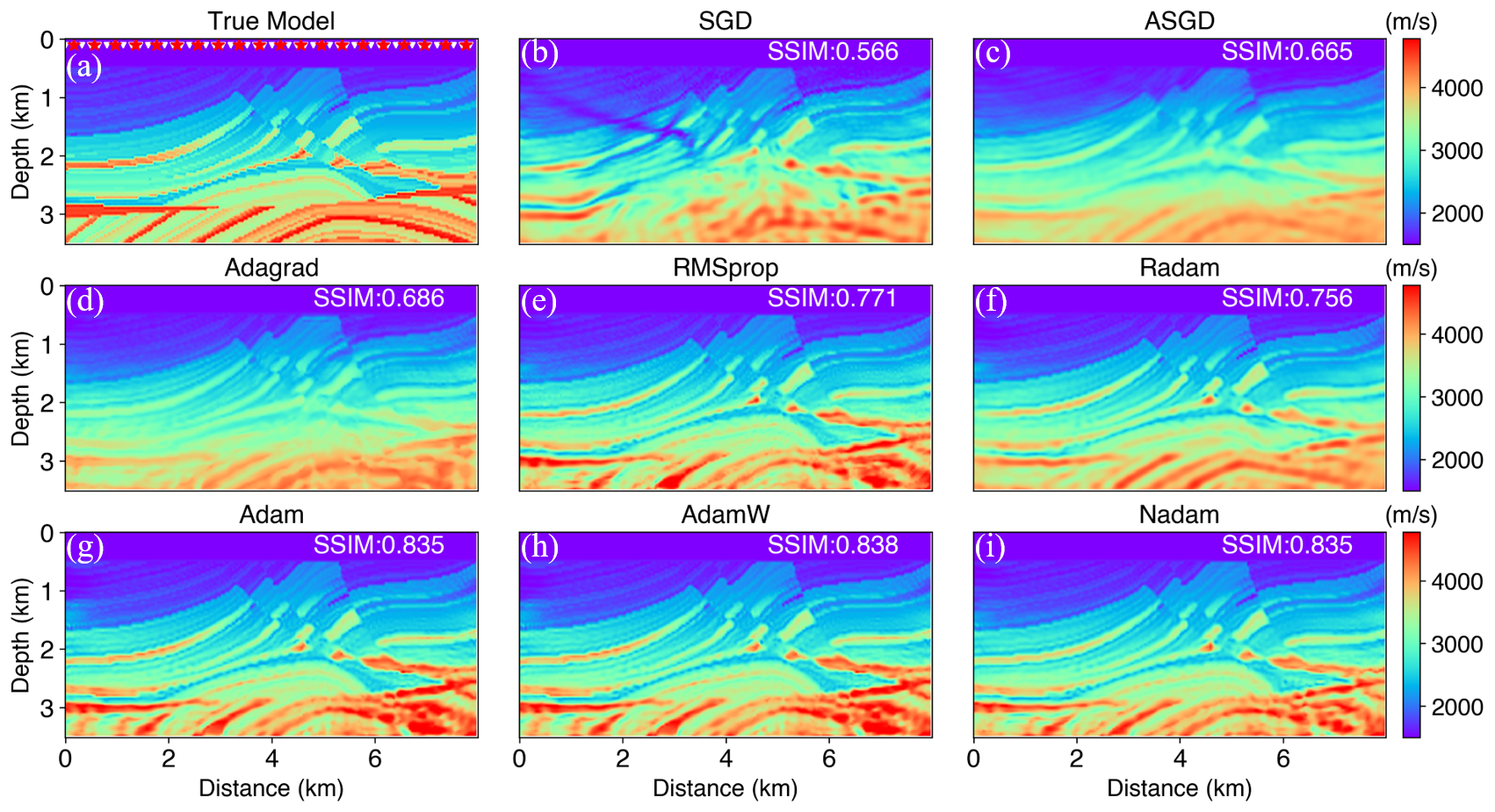}
                \caption{Comparison for eight different optimizers for ADFWI. (a) The true velocity model, with red stars indicating the source locations and white triangles indicating the receiver positions. Inversion results using the (b) SGD, (c) ASGD, (d) Adagrad, (e) RMSprop, (f) Radam, (g) Adam, (h) AdamW, and (i) NAdam optimization methods. The initial model is obtained by applying a 240 $\times$ 240 m Gaussian smoothing filter to the true model.}
                \label{Figure8}
            \end{figure}

            \begin{figure}[!httbp]
                \centering
                \includegraphics[width=0.9\textwidth]{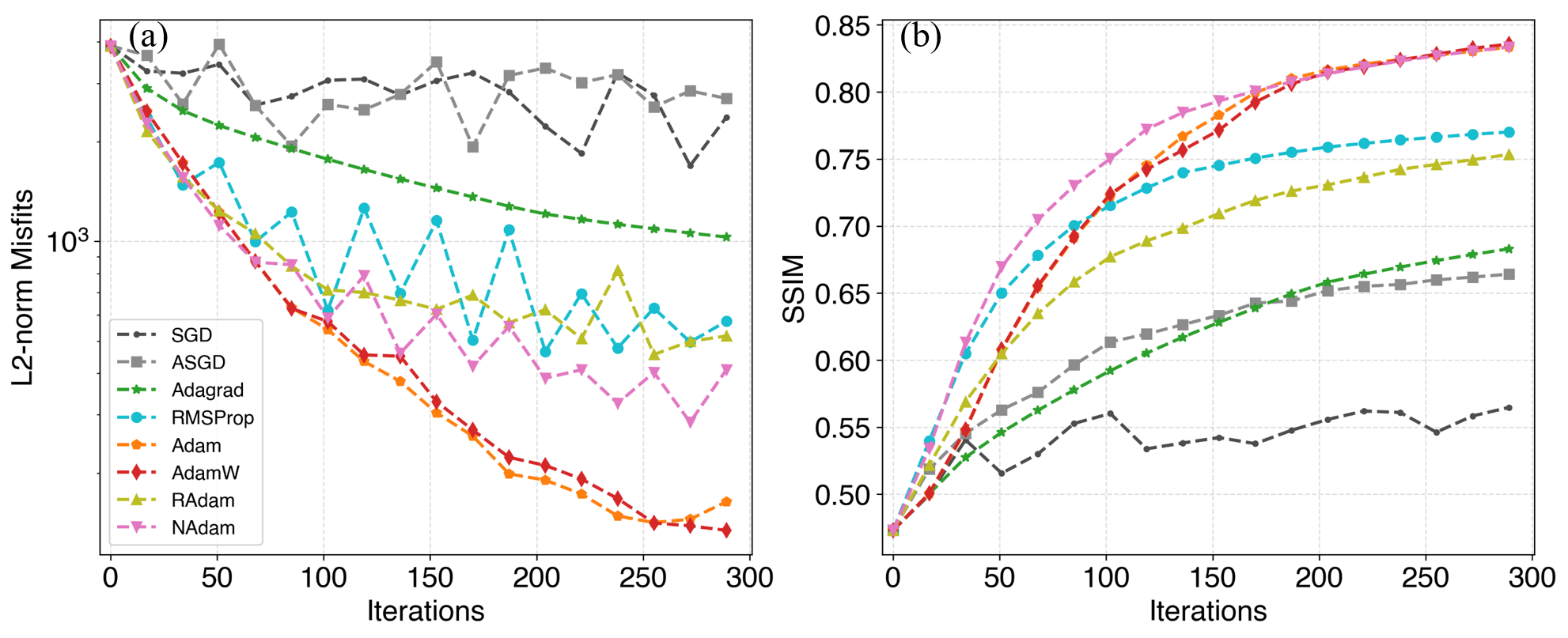}
                \caption{Variations of data misfit and model similarity (SSIM) for different optimizers. (a) The L2-norm for data misfit, and (b) the SSIM for model similarity for eight different optimizers with iterations.}
                \label{Figure9}
            \end{figure}
            
            \begin{table}[!httbp]
                \caption{Comparison of MAPE and SSIM for the true and inverted models with different optimization methods. The best results are highlighted in bold. For MAPE, lower values indicate better models, while for SSIM higher values indicate better models.}
                \resizebox{\textwidth}{!}{%
                \begin{tabular}{c|cccccccc}
                    \hline
                        \diagbox{Metric}{Optimizers} & SGD & ASGD & Adagrad & RMSProp & RAdam & Adam & NAdam & Adamw \\ \hline
                        MAPE & 7.954 & 6.559 & 6.332 & 5.842 & 5.719 & 4.981 & \textbf{4.865} & 4.927 \\
                        SSIM & 0.566 & 0.665 & 0.686 & 0.771 & 0.756 & 0.835 & 0.835 & \textbf{0.838} \\ 
                    \hline
                \end{tabular}
                }
            \end{table}

    \subsection{Evaluation of Conventional and Neural Network-based Regularization in ADFWI}

        FWI is inherently a non-unique problem, particularly when field data are contaminated with noise, acquisitions are limited in frequency contents, illumination, or azimuth coverages, or wave equations cannot genuinely honor seismic propagation in the real Earth \cite{virieux_2009_Overview, fichtner_2011_Resolution}. Regularization that integrates prior information and penalizes drastic changes can help stabilize inversion and yield more plausible results. In addition to traditional regularization methods, ADFWI can also conveniently integrate DNN for model constraints \cite{he_2021_Reparameterized, zhu_2022_Integrating, sun_2023_FullWaveform, wang_2023_Prior}.
        
        \subsubsection{Integration of Regularization through Automatic Differentiation}
            
            In this section, we evaluate conventional regularization in ADFWI using the ISO-acoustic Marmousi2 model, following the same test configuration outlined in Section 3.1.1. Independent Gaussian noises are added to each individual trace, with the mean value equal to that of each trace and standard deviation four times the standard deviation of the raw waveforms (Figure 10a), yielding a mean signal-to-noise ratio of -4.10 dB. The global correlation-based objective function is used to better suppress influence from noise \cite{choi_2012_Application, tao_2017_Fullwaveform}, and the inversion is iterated 300 times with the Adam optimizer. It is clear that the inverted model without regularization contains severe artifacts (Figure 10b), and in comparison, the models with first- and second-order Tikhonov and TV regularization are better recovered (Figures 10c-f). Detailed metrics for the regularization is provided in Table 4. Figure 11 shows the change in data misfit and model similarities with iterations. 
            
            Though implementations of the Tikhonov and TV regularization are relatively simple, more sophisticated regularization methods using seislet \cite{xue_2017_Fullwaveform}, dictionary learning \cite{li_2018_Full}, and diffusion model \cite{wang_2023_Prior, taufik_2024_Learned} could be quite involved to use for traditional FWI. For instance, the regularization with seislet requires transforming the model into the seislet domain, where the gradient computation is intricately related to the selected basis function and the transformation process \cite{xue_2017_Fullwaveform}. In comparison, the new AD-based framework largely simplifies the implementation by fully automating gradient computations.
            
            \begin{figure}[!ht]
                \centering
                \includegraphics[width=0.85\textwidth]{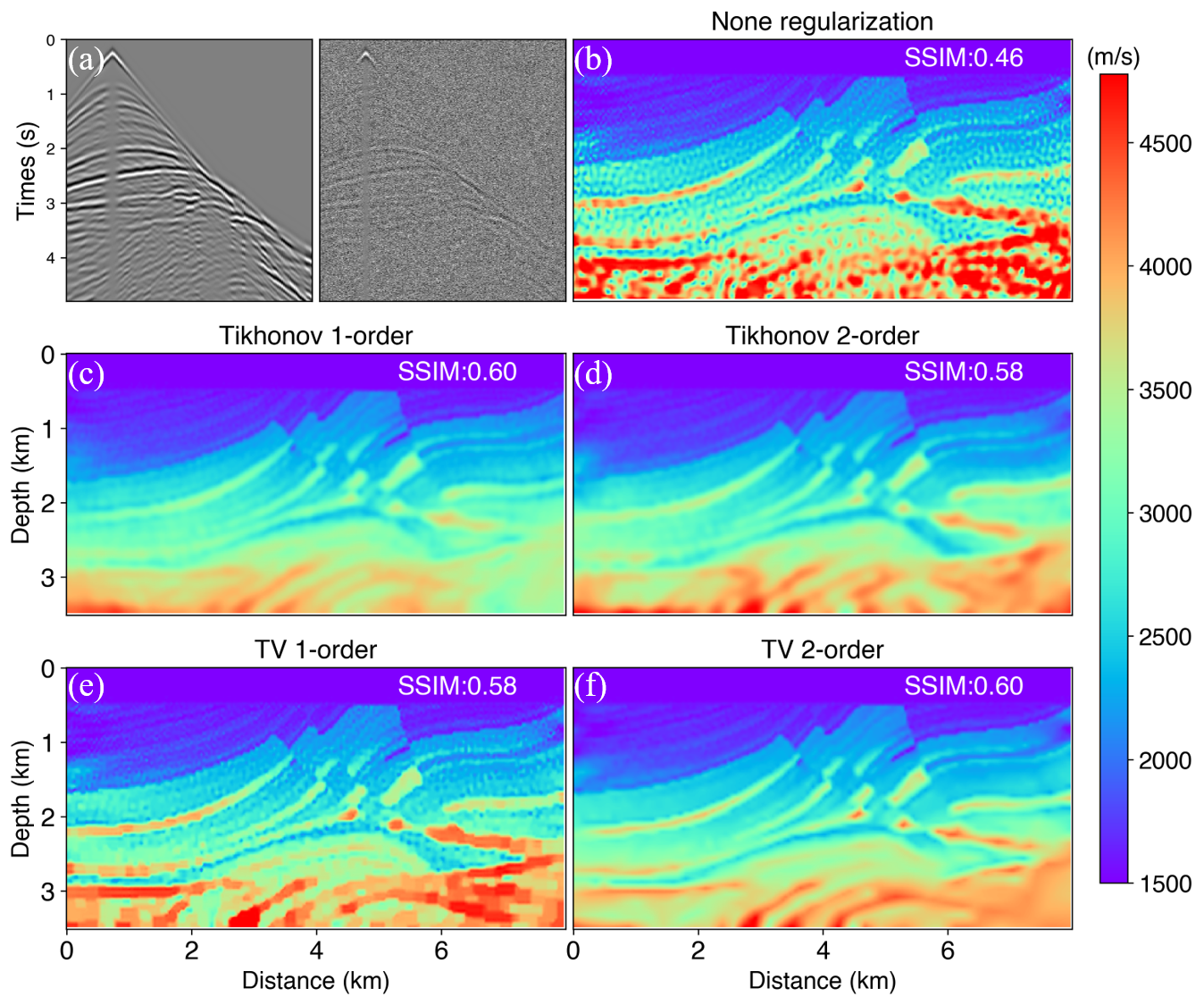}
                \caption{Comparison of the inversion results using different regularization methods. (a) The raw dataset (left panel) and noisy dataset which has a mean signal-to-noise ratio of -4.10 dB (right panel) used for the tests; (b) inverted model without regularization; (c)-(d) inverted models using the first- and second-order Tikhonov regularization; (e)-(f) inverted models using the first- and second-order total variation (TV) regularization.}
                \label{Figure10}
            \end{figure}

            \begin{figure}[!ht]
                \centering
                \includegraphics[width=0.85\textwidth]{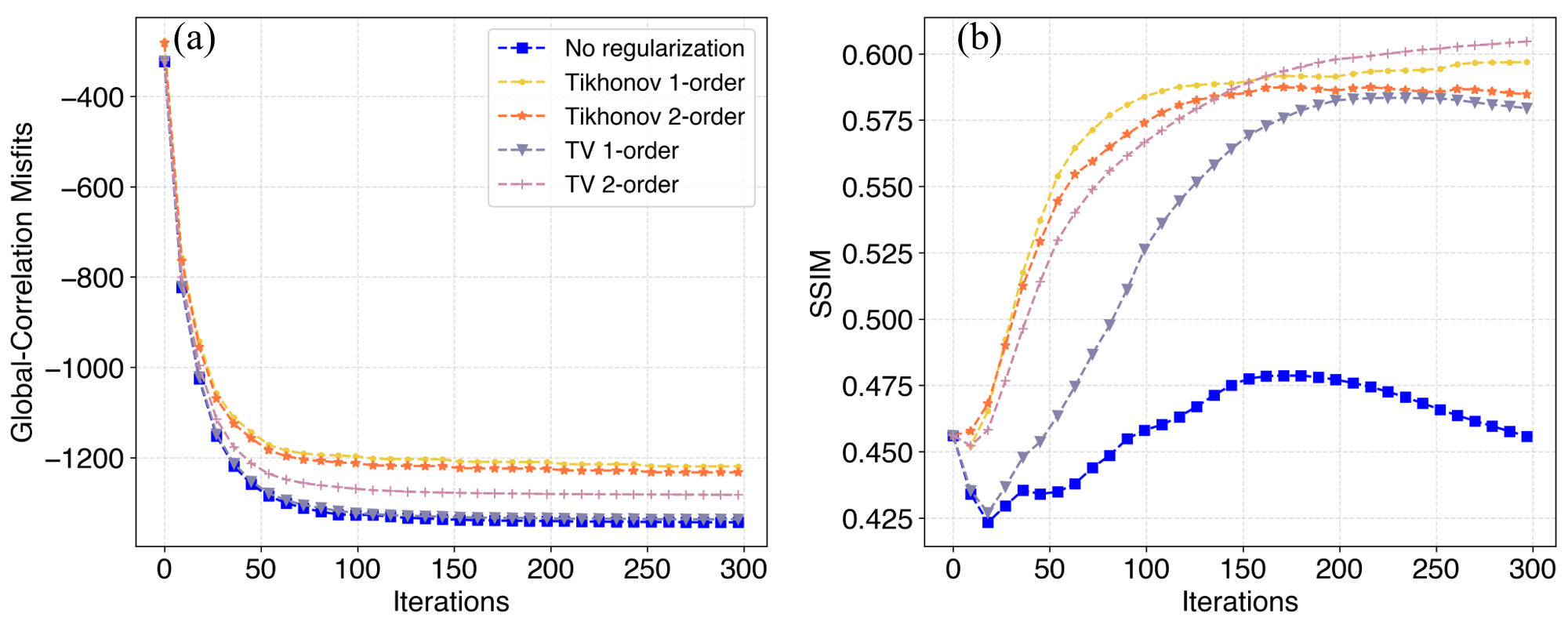}
                \caption{Variations of data misfit and model similarity (SSIM) with iterations using the global correlation objective function with various regularization terms. (a) Change of data misfit, and (b) change of SSIM with iterations.}
                \label{Figure11}
            \end{figure}
            
            \begin{table}[!ht]
                \caption{Comparison of MAPE and SSIM for different regularization methods. The best results are highlighted in bold.}
                \resizebox{\textwidth}{!}{%
                \begin{tabular}{c|ccccc}
                    \hline
                        \diagbox{Metric}{Regular.} & No Regularization & Tikhonov 1st-order& Tikhonov 2nd-order& TV 1st-order& TV 2nd-order\\ \hline
                        MAPE & 8.504 & 7.242 & 6.901 & 6.648 & \textbf{6.409} \\
                        SSIM & 0.455 & 0.597 & 0.585 & 0.579 & \textbf{0.605} \\ 
                    \hline
                \end{tabular}
                }
            \end{table}
        
        \subsubsection{Learned Regularization based on Deep Neural Networks}
            
            Reparameterization of the model using DNNs can introduce implicit and learned regularization constraints in the inversion results \cite{he_2021_Reparameterized, zhu_2022_Integrating}. In this section, we further assess the effectiveness of reparameterization using the ISO-acoustic Marmousi2 model, and the test configuration is described in Section 3.3.1. We use the CNN-based neural network for reparameterization with varying numbers of blocks (Figure 2), and the input is a vector with 100 random elements. Prior to performing FWI, a pre-training strategy is first used to learn the end-to-end mapping between random input vectors and a known initial velocity model derived from, e.g., travel-time inversion. Subsequently, ADFWI iteratively updates the neural network parameters which in return progressively refine the output model. As shown in Figure 12, the inversion results vary distinctly with the number of CNN blocks. Reparameterization with fewer blocks can effectively suppress noise in the models while preserving sharper boundaries, and reparameterization with an increasing number of CNN blocks leads to smoother velocity models in comparison. Thus, DNN should be tailored in practice to balance data fitting against model smoothness by adjusting the network architecture. Also, recent studies indicate that combining DNNs with conventional regularization techniques such as TV or Tikhonov can lead to further improved results. It should be noted while in this study we only used multilayer CNN for reparameterization as a proof of concept, alternative neural networks could yield even superior performance, which can be further explored with ADFWI by interested readers. More details for the network structure can be found in Supporting Table S1. The metrics for this comparison are detailed in Table 5, and the extracted velocity profiles from the models inverted with both traditional and DNN-based regularization are shown in Figure S7. 
            
            \begin{figure}[!ht]
                \centering
                \includegraphics[width=0.9\textwidth]{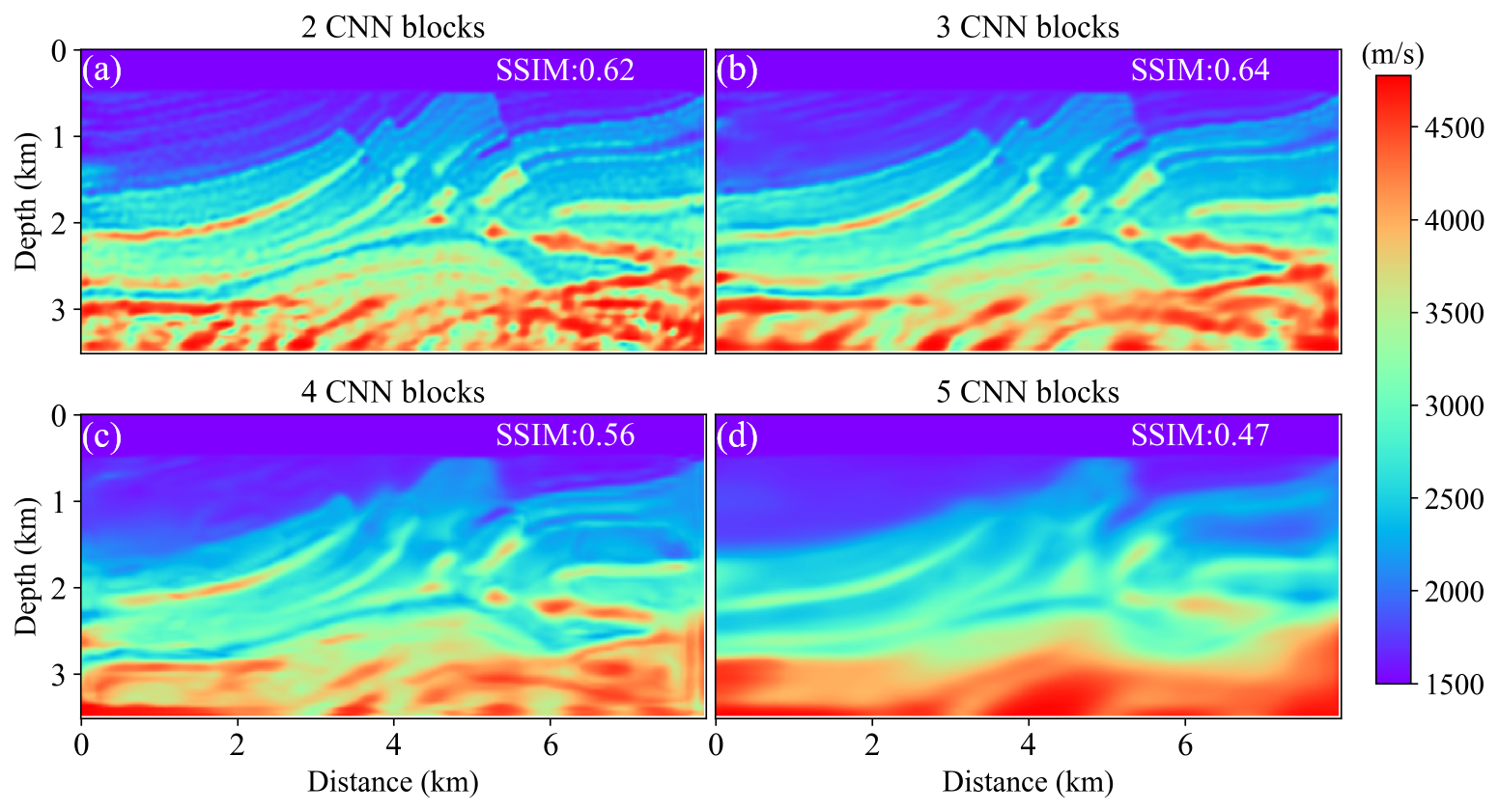}
                \caption{Comparison of the inversion results using neural networks with different numbers of CNN blocks. (a) The inverted model with 2 CNN blocks, (b) 3 CNN blocks, (c) 4 CNN blocks and (d) 5 CNN blocks for model reparameterization. The true and initial model used for FWI are shown in Figures 3a and 3b}
                \label{Figure12}
            \end{figure}

            \begin{table}[htb]
            \caption{Comparison of MAPE, SSIM, and the sizes of the neural networks for reparameterization with different numbers of CNN blocks. The best results are highlighted in bold.}
            \resizebox{\textwidth}{!}{%
                \begin{tabular}{c|ccccc}
                    \hline
                    \diagbox{Metrics}{Regularization} & No Regularization & 2 CNN blocks & 3 CNN blocks & 4 CNN blocks & 5 CNN blocks \\ \hline
                    MAPE & 8.504 & 5.819 & \textbf{5.544} & 6.438 & 9.624 \\
                    SSIM & 0.455 & 0.628 & \textbf{0.637} & 0.582 & 0.457 \\
                    Model Size & 17,600& 1,762,560 & 1,776,640 & 1,805,312 & 2,169,088 \\ \hline
                \end{tabular}
                }
            \end{table}

%
%
\section{Discussion}
    \subsection{Customized Hybrid Objective Functions based on the AD Library}

        As discussed above, ADFWI greatly simplifies gradient calculations associated with different objective functions and regularization schemes. Therefore, different objective functions such as the envelope and global correlation can be assembled to create new hybrid functions that leverage respective strengths \cite{vyas_2022_FWI}. Figure 13 shows the inversion results using such a hybrid objective function, namely the weighted envelope correlation-based objective function \cite{song_2023_Weighted}. In the inversion process, a larger weight is initially assigned to the envelope objective function to reduce local minima, and the weight for the global correlation objective function is gradually increased with iterations to better recover fine structures. The results by individual objective functions are also provided for comparison. Detailed information for the hybrid objective function and its implementation are provided in Supporting Text S1.

        \begin{figure}[!ht]
            \centering
            \includegraphics[width=1.0\textwidth]{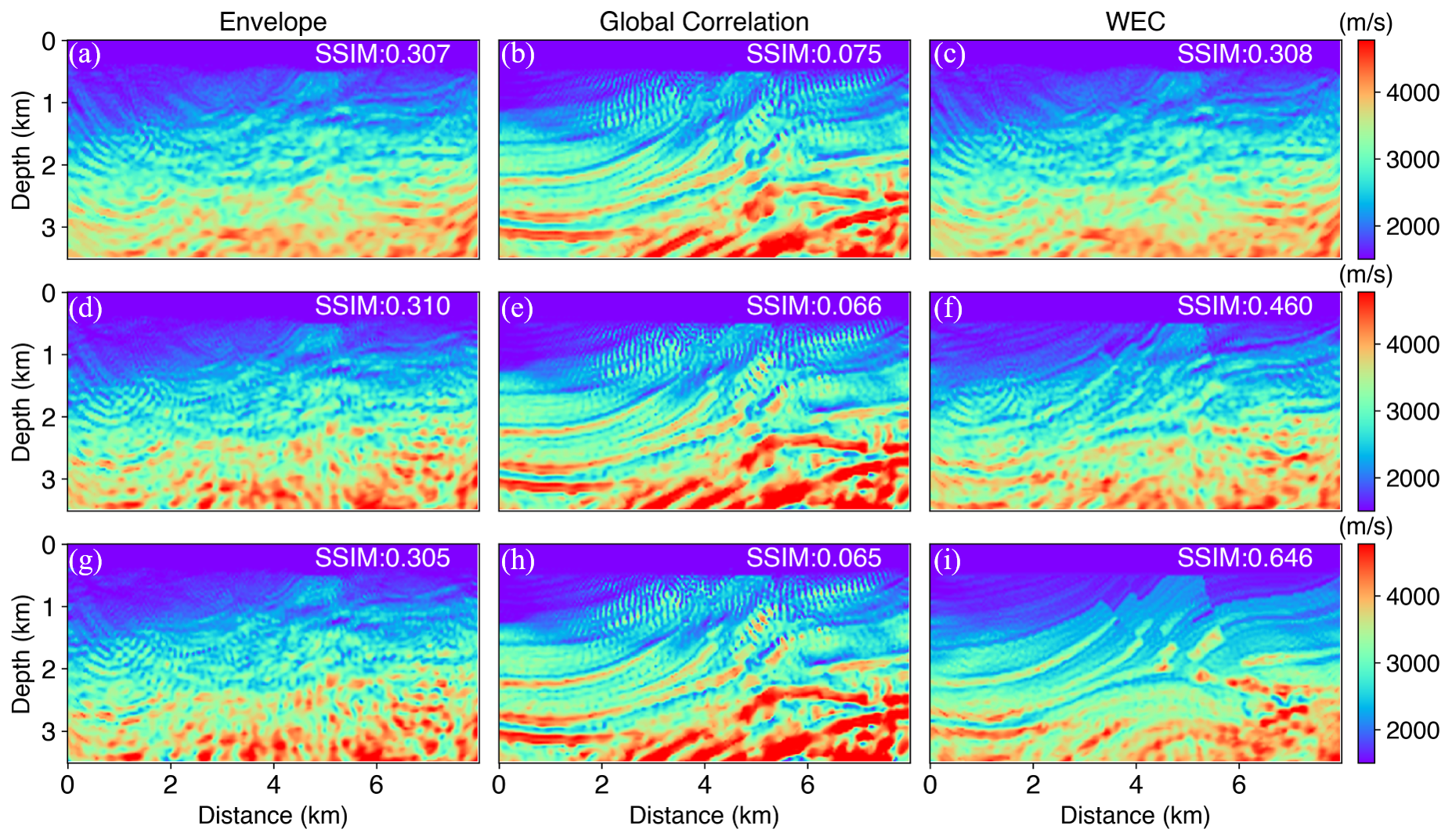}
            \caption{Comparison of the inversion results using individual and hybrid objective functions. The inversion results obtained by the (a) envelop, (b) global correlation, and (c) WEC hybrid objective functions at the 100th iteration; (d)-(f) are similar to (a)-(c), but show the results at the 200th iterations; (g)-(i) show the results at the 300th iterations.}
            \label{Figure13}
        \end{figure}
        
    \subsection{Uncertainty Assessment using Dropout in Deep Neural Network}
    
        By reparameterizing the velocity model with DNN, not only we can incorporate learned regularization to enhance stability of FWI, but specific neural network techniques such as dropout can be used to estimate local uncertainties in FWI \cite{sun_2021_Physicsguided, zhu_2022_Integrating}. Dropout, which was originally developed to prevent overfitting by randomly deactivating a portion of neurons and their connections in training, can be used to approximate the Bayesian posterior distribution of inverted models by inferencing multiple times through partly deactivating the trained neural network \cite{srivastava_2014_Dropout, gal_2015_Dropout}. In ADFWI, once the neural network for reparameterization is optimized, a great number of velocity models can be rapidly generated by applying dropout during inference, and the corresponding standard deviation of the sampled velocity models can be used as an estimate for the model uncertainty. In this study, we perform inference 1,000 times using the optimized 2-layer CNN, which is described in the learned regularization tests in Section 3.3.2. Figure 14b shows the inversion result without dropout (same as Figure 12a), and Figures 14c and 14d show the corresponding estimated uncertainties with dropout rate \(p\) of 0.1 and 0.2. Here \(p\) represents the proportion of neurons that are randomly deactivated during each forward pass, and higher \(p\) is related to larger model perturbations. The estimated model uncertainty is more significant at greater depths and along structural boundaries, which can be explained by weakly reflected waveforms from deep layers and smearing due to regularization, respectively. Thus, dropout in ADFWI can provide an efficient means to assess the local uncertainty in the results compared to the computationally demanding Bayesian approach \cite{gal_2015_Dropout}.
        
        \begin{figure}[!ht]
            \centering
            \includegraphics[width=0.9\textwidth]{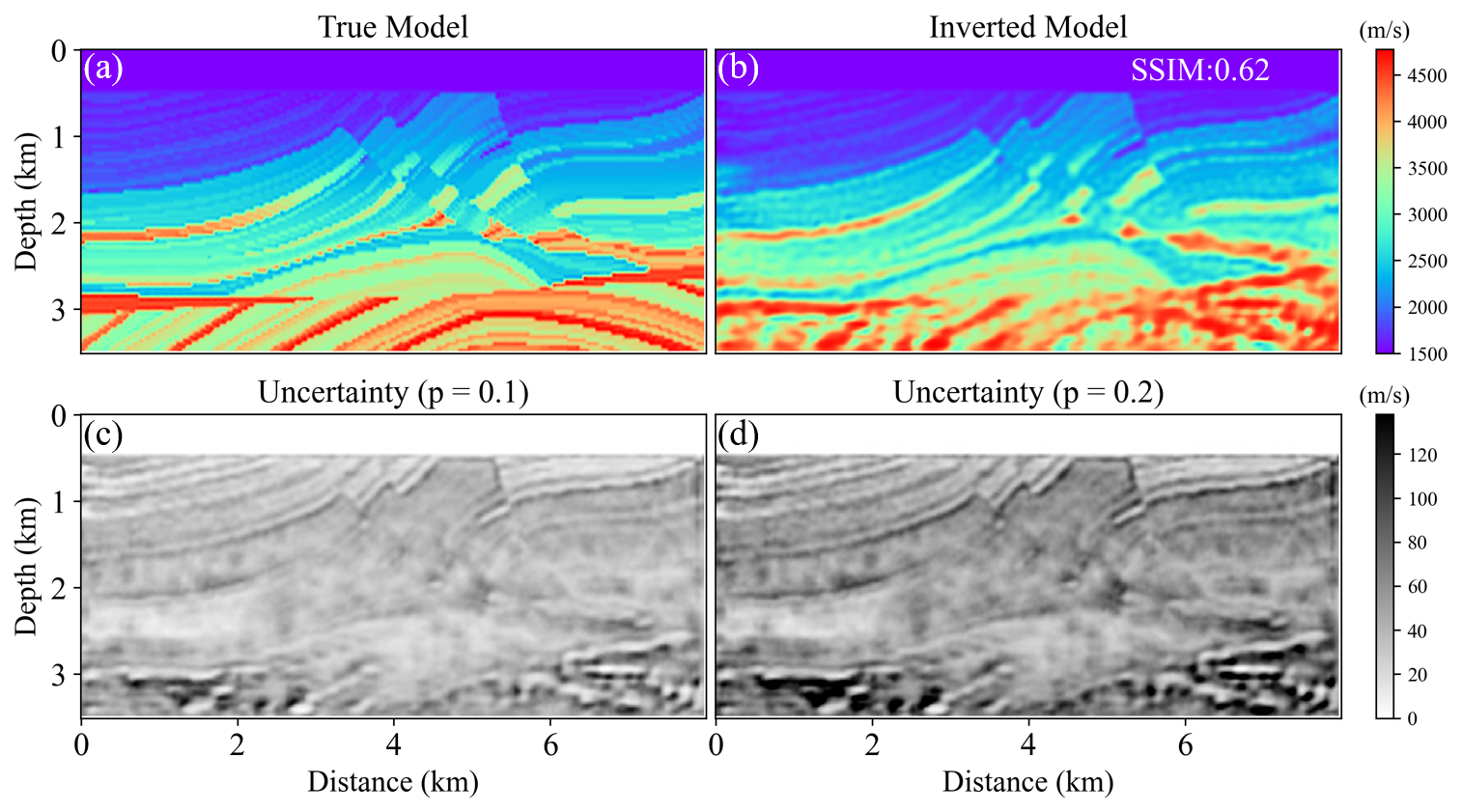}
            \caption{Uncertainty estimation using the two-layer CNN optimized in the learned regularization tests outlined in Section 3.3.2. (a) The true Marmousi2 model; (b) the inverted model with no neurons deactivated, same as Figure 12a; estimated uncertainty based on 1,000 models inferred from the two-layer CNN with dropout rate (c) \( p \)=0.1, and (d) \( p \)=0.2.}
            \label{Figure14}
        \end{figure}

    \subsection{Computational Consumption and Reduction Strategies}
        The use of AD in FWI is computationally demanding, particularly in memory overhead, which is due to the storage of intermediate gradients throughout the computational graph \cite{baydin_2015_Automatic, wang_2023_Memory}. We use two strategies, including the mini-batch processing and checkpointing to tackle these challenges. The first strategy involves segregating seismic datasets into smaller subsets or mini-batches \cite{krizhevsky_2012_ImageNet, kazei2023acquisition}, which are back-propagated sequentially. The mini-batch processing can reduce memory consumption in the forward modeling by only calculating gradients from part of the sources, which allows performing FWI for large-scale problems with reasonable computational resources, albeit at the cost of more iterations. Note we first accumulate the gradients from all individual mini-batched sources before updating the model. 
        
        Checkpointing further reduces memory consumption in gradient calculation with the computational graph by saving only selected intermediate states in forward modeling, segmenting the process into different time intervals, and storing only critical information at specified checkpoints \cite{gruslys_2016_Memoryefficient}. In back-propagation, some temporarily discarded intermediate states (i.e., the wavefield) in the forward pass are recalculated, which can significantly reduce the overall memory usage at the cost of slightly increased computational time. These two strategies combined, can strike a balance between computational efficiency and memory consumption, and enable practical applications of AD-based FWI on large datasets or complex models. Figure 15 shows the impact of the two strategies on memory demand and computational time based on the ISO-acoustic Marmousi2 model described in Section 3.1.1 on a single NVIDIA GeForce RTX 4090 GPU. It is evident that there is a tradeoff between memory consumption and computational efficiency. However, the computational time is increased slightly with larger checkpoint segments, while the memory consumption is reduced significantly (Figure 15a). On the other hand, the batch size (i.e., number of shots inverted simultaneously) is linearly proportional to the memory consumption, while the computational time for all shots in one iteration is reversely proportional to the batch size. Therefore, though the optimal batch size and checkpoint segment can vary with a particular inverse problem, it can be recommended in general that a larger batch size and a smaller checkpoint segment should be opted for as long as the GPU memory can accommodate. 

        \begin{figure}[!ht]
            \centering
            \includegraphics[width=1.0\textwidth]{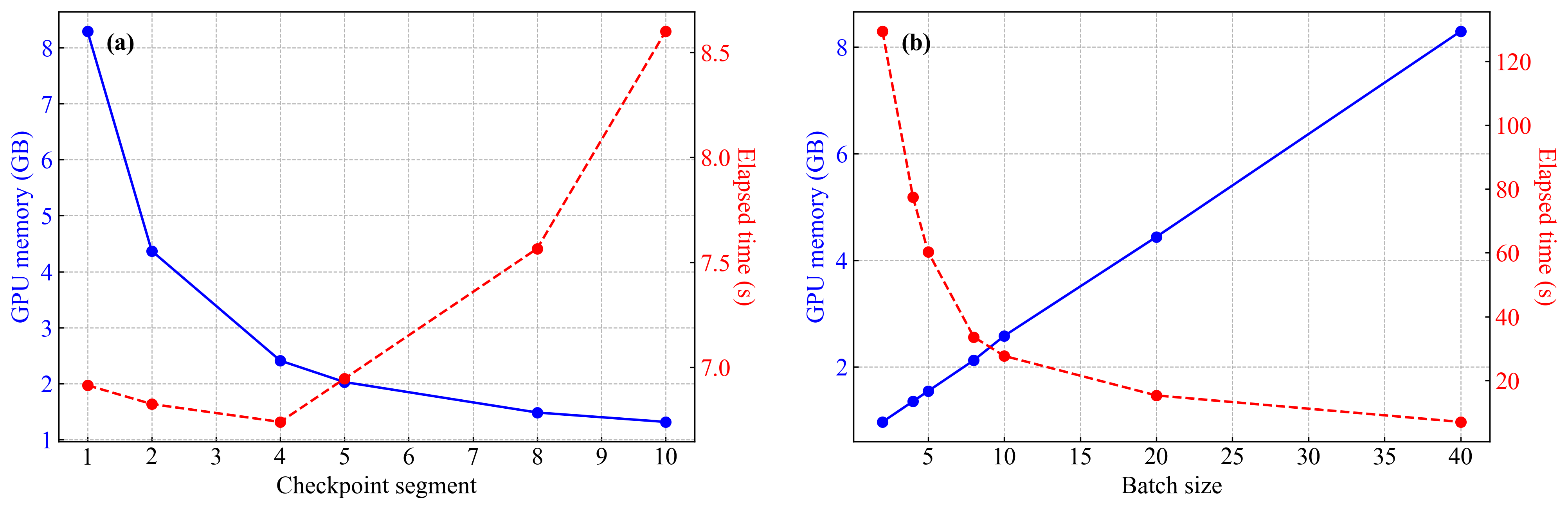}
            \caption{Comparison of memory demand (blue) and runtime per iteration (red) with different checkpointing and mini-batch strategies. The tests are based on the ISO-acoustic Marmousi2 model described in Section 3.1.1 on a single NVIDIA GeForce RTX 4090 GPU. (a) Impact of varying checkpoint segments on memory and runtime while maintaining a fixed batch size of 40, i.e., all shots are inverted simultaneously. The number of checkpoint segments means how many segments in time the entire forward modeling process is divided into. (b) Impact of different batch sizes on memory and runtime while fixing the number of checkpoint segments to 1.}
            \label{Figure15}
        \end{figure}

%
%
\section{Conclusion}
    In this study, we present ADFWI, an automatic differentiation-based, open-source framework for full waveform inversion. Taking advantage of the well-implemented automatic differentiation in modern deep learning libraries such as Pytorch, ADFWI eliminates the need for manual derivation and implementation of adjoint equations and sources, providing a unified yet flexible platform for waveform modeling and inversion in ISO-acoustic, ISO-elastic, and VTI/HTI elastic media. The framework also provides a variety of objective functions, optimization methods, as well as traditional and neural network-based regularization techniques for choice. Local uncertainty in the inverted models can also be rapidly estimated through dropout when the model is reparameterized with a neural network. The architecture of ADFWI allows easy integration of additional types of wave equations and inversion strategies not considered in the current platform. 
 
    Comprehensive synthetic tests have been conducted to validate all the components in ADFWI. Using the mini-batch processing and checkpointing strategies, the high memory requirement in automatic differentiation is considerably reduced. It is hoped that ADFWI will become a useful workbench for seismologists not only to explore new ideas in waveform inversion, but also to better characterize the Earth's complex subsurface structures.

%
%
\section*{Code and Data Availability Statement}
    The open-source automatic differentiation-based full waveform inversion package ADFWI developed in this study is available at \url{https://github.com/liufeng2317/ADFWI}. All tests performed in this study can be reproduced using scripts provided in the examples folder. The Marmousi2 model used in the study is available at \url{http://www.ahay.org/RSF/book/data/marmousi2/paper_html}, and the Overthrust model can be found at \url{https://s3.amazonaws.com/open.source.geoscience/open_data/seg_eage_models_cd/salt_and_overthrust_models.tar.gz}. 

%
%
\section*{Acknowledgments}
    This research is supported by the Joint Funds of National Natural Science Foundation of China (U2139204). Computational resources provided by the Shanghai Artificial Intelligence Laboratory are also greatly appreciated.

%
%
\bibliography{reference}

\end{document}


\justifying
%
%
\begin{center}
    \Large \textbf{Supporting Information for \\ "Automatic Differentiation-based Full Waveform Inversion with Flexible Workflows"}
\end{center}

\begin{center}
    Feng Liu$^{1,2}$, Haipeng Li$^{2,3*}$, Guangyuan Zou$^{2}$, Junlun Li$^{2,4,5*}$
\end{center}

\begin{center}
    \small
    $^1$ School of Electronic Information and Electrical Engineering, Shanghai Jiao Tong University, Shanghai 200240, China.\\
    $^2$ Laboratory of Seismology and Physics of Earth’s Interior, School of Earth and Space Sciences, University of Science and Technology of China, 96 Jinzhai Road, Hefei 230026 Anhui, China.\\
    $^3$ Now at Department of Geophysics, Stanford University, CA 94305, USA.\\
    $^4$ Mengcheng National Geophysical Observatory, University of Science and Technology of China, Hefei 230026 Anhui, China.\\
    $^5$ CAS Center for Excellence in Comparative Planetology, 96 Jinzhai Road, Hefei 230026 Anhui, China.
\end{center}

\hspace{10pt}

{\small
\RaggedRight
* Corresponding authors: Haipeng Li (haipeng@sep.stanford.edu), Junlun Li (lijunlun@ustc.edu.cn)
}

\vspace{15pt}
%
%
%
%
\noindent\textbf{Contents of this file}
\begin{enumerate}
    \item Text S1 to S2.
    \item Table S1.
    \item Figures S1 to S7.
\end{enumerate}

\noindent\textbf{Introduction}

    This supplementary file provides descriptions and derivations of the various objective functions and optimization algorithms integrated in the automatic differentiation-based full waveform inversion framework (ADFWI). Additionally, Table S1 provides a detailed description of the architecture of the CNN-based neural network for reparameterization and learned regularization. Furthermore, Figure S1 shows the validation of the accuracy of AD in gradient computation. Figure S2 shows an additional test for simultaneous multi-parameter inversion of $v_p$ and $\rho$. Figure S3 presents a comparison of the convexity of various objective functions using a time-shifted Ricker wavelet. Figures S4, S5 and S7 show the extracted P-wave velocity profiles from velocity models inverted by different objective functions, optimization algorithms and regularization methods, respectively. Figure S6 shows the comparison of the inversion results using the \textit{l}-BFGS and AdamW optimization methods. Together, these contents provide additional evidences to support findings in the main text.  

\clearpage

%
%
\noindent\textbf{Text S1: Objective Functions}

\vspace{15pt}
\noindent\textbf{(1) L2-norm-based objective function}
    \begin{equation}
        \mathcal{J}_{L2}(\mathbf{m}) = \frac{1}{2} \sum_{s} \sum_{g} \int_{0}^{T} ||d_{obs}(s, r, t) - d_{cal}(\mathbf{m}; s, r, t)||^2 dt,
    \end{equation}
    where $d_{obs}(s, r, t)$ and $d_{cal}(\mathbf{m}; s, r, t)$ represent the observed and synthetic data, respectively, for each shot ($s$) and receiver ($r$); $t$ denotes the recording time, $T$ denotes the maximum recording time, and $\mathbf{m}$ denotes the model parameters. 

\vspace{15pt}
\noindent\textbf{(2) L1-norm-based objective function}
    \begin{equation}
        \mathcal{J}_{L1}(\mathbf{m}) = \sum_{s} \sum_{g} \int_{0}^{T} |d_{obs}(s, r, t) - d_{cal}(\mathbf{m}; s, r, t)| dt,
    \end{equation}
    where the definitions are the same as those for $\mathcal{J}_{L2}(\mathbf{m})$.

    

\vspace{15pt}
\noindent\textbf{(3) T-distribution-based objective function}
    \begin{equation}
        \mathcal{J}_{StudentT}(\mathbf{m}) = \sum_{s} \sum_{g} \int_{0}^{T} \frac{n+1}{2} \log\left[1 + \frac{1}{n \sigma^2} ||d_{obs}(s, r, t) - d_{cal}(\mathbf{m}; s, r, t)||^2\right] dt,
    \end{equation}
    where $\sigma$ is the scaling parameter and $n$ denotes the degrees of freedom of the T-distribution (Aravkin et al., 2011; Guo et al., 2023).

\vspace{15pt}
\noindent\textbf{(4) Envelope-based objective function}
    \begin{equation}
        \mathcal{J}_{Envelope}(\mathbf{m}) = \sum_{s} \sum_{g} \int_{0}^{T} ||E_{obs}^p(s, r, t) - E_{cal}^p(\mathbf{m}; s, r, t)||^2 dt,
    \end{equation}
    where $E_{obs}(t) = \sqrt{d_{obs}^2(t) + \widetilde{d}_{obs}^2(t)}$ is the envelope of $d_{obs}(t)$, and $\widetilde{d}_{obs}(t)$ is the Hilbert transform of $d_{obs}(t)$; $E_{syn}(t) $ is similar to $E_{obs}(t) $ but for synthetic data; $p$ represents an operation on $E(t)$, such as absolute values ($p=1$) or squares ($p=2$) (Bozda{\u{g}} et al., 2011; Wu et al., 2014).

\vspace{15pt}
\noindent\textbf{(5) Global correlation-based objective function}

    The zero-lag cross-correlation between two normalized waveforms can be defined as (Choi \& Alkhalifah, 2012):
    \begin{equation}
        \mathcal{J}_{GC}(\mathbf{m}) = \sum_{s} \sum_{r} \int_{0}^{T} \left[ 1 - \hat{d}_{obs}(s, r, t) \cdot \hat{d}_{cal}(\mathbf{m}; s, r, t) \right] dt,
    \end{equation}
    where $\hat{d}_{obs}(s, r, t) = d_{obs}(s, r, t)/\left\| d_{obs}(s, r, t) \right\|$ and $\hat{d}_{cal}(\mathbf{m}; s, r, t) = d_{cal}(\mathbf{m}; s, r, t) / \left\| d_{cal}(\mathbf{m}; s, r, t) \right\|$.

\clearpage
\noindent\textbf{(6) Weighted envelope correlation-based objective function}

    The weighted envelope correlation-based objective function (WEC) combines the advantages of the global correlation and the envelope objective functions (Song et al., 2023):
    \begin{equation}
        \mathcal{J}_{WEC}(\mathbf{m}) = w(i) \mathcal{J}_{GC}(\mathbf{m}) + (1 - w(i))\mathcal{J}_{Envelope}(\mathbf{m}),
    \end{equation}
    where $w(i)$ denotes a weighting factor, with $i$ representing the iteration number of the inversion. We use the sigmoid function to define the weighting factor $w(i)$:
    \begin{equation}
        w(i) = \frac{1}{1 + e^{-i - \frac{N}{2}}}, \quad \text{for } i = 1, 2, \ldots, N,
    \end{equation}
    where $N$ is the number of iterations.

\vspace{15pt}
\noindent\textbf{(7) Dynamic time warping-based objective function}

    Dynamic time warping (DTW) is a method used to measure the similarity between two time sequences. Given two time series $X = [x_1, x_2, \ldots, x_N] $ and $Y = [y_1, y_2, \ldots, y_M]$, the DTW distance $D(X, Y)$ is defined as:
    \begin{equation}
        D(X, Y) = \min_{W} \left( \sum_{(i, j) \in W} d(x_i, y_j) \right),
    \end{equation}
    where $W$ represents a warping path from $(1,1)$ to$(N,M)$, and $d(x_i, y_j)$ denotes the distance between points $x_i$ and$y_j$ (typically the Euclidean distance). The cumulative distance matrix is updated using:
    \begin{equation}
        D(i, j) = d(x_i, y_j) + \min \{ D(i-1, j), D(i, j-1), D(i-1, j-1) \}.
    \end{equation}
    
    In FWI, we calculate the DTW distance between the observed data $d_{obs}(s, r, t)$ and synthetic data $d_{cal}(\mathbf{m}; s, r, t)$ for each source-receiver pair $(s, r)$. The DTW-based objective function is then defined as (Ma \& Hale, 2013; Cuturi \& Blondel, 2017; Chen et al., 2022):
    \begin{equation}
        \mathcal{J}_{DTW}(\mathbf{m}) = \sum_{s} \sum_{r} D(d_{obs}(s, r, t), d_{cal}(\mathbf{m}; s, r, t)).
    \end{equation}
    
    The details of the calculations are as follows:
    \begin{enumerate}
        \item \textbf{Initialization:} Construct the cumulative distance matrix $D$ with size $(N+1) \times (M+1)$ and initialize all elements to infinity, except $D(0, 0) = 0$.
        \item \textbf{Update cumulative distance matrix:} For each $(i, j)$, update the matrix using:
        \begin{equation}
            D(i, j) = d(d_{obs}(s, r, t_i), d_{cal}(\mathbf{m}; s, r, t_j)) + \min \{ D(i-1, j), D(i, j-1), D(i-1, j-1) \}.
        \end{equation}
        \item \textbf{Find the optimal path:} The optimal path is traced back from $(N, M)$ to $(1, 1)$, accumulating the distances along the path.
    \end{enumerate}
    
    This DTW-based objective function can handle time shifts and distortions between the observed and simulated data, providing a more robust and flexible measurement of misfits compared to the traditional waveform-difference-based objective functions.

\vspace{15pt}

\noindent\textbf{(8) Wasserstein distance-based objective function}

    The Wasserstein-Sinkhorn distance is a measurement used to quantify the difference between two probability distributions. It combines the Wasserstein distance with the Sinkhorn regularization to balance computational efficiency and stability. Consider two discrete probability distributions $\mu$ and $\nu$:
    \begin{equation}
        \mu = \sum_{i=1}^{n} \mu_i \delta_{x_i}, \quad \nu = \sum_{j=1}^{m} \nu_j \delta_{y_j},
    \end{equation}
    where $\delta_{x_i}$ and $\delta_{y_j}$ are Dirac functions at $x_i$ and $y_j$, respectively, and $\mu_i$ and $\nu_j$ are the weights (probability masses) at these locations. The Wasserstein distance is defined as:
    \begin{equation}
        W(\mu, \nu) = \min_{\gamma \in \Gamma(\mu, \nu)} \sum_{i=1}^{n} \sum_{j=1}^{m} \gamma_{ij} d(x_i, y_j),
    \end{equation}
    where $\Gamma(\mu, \nu)$ denotes the set of all joint probability distributions satisfying the marginal constraints:
    \begin{equation}
        \sum_{j=1}^{m} \gamma_{ij} = \mu_i, \quad \sum_{i=1}^{n} \gamma_{ij} = \nu_j.
    \end{equation}
    
    Sinkhorn regularization introduces an entropy regularization term to make the computation more efficient. With a regularization parameter $\lambda > 0$, the Sinkhorn distance is defined as:
    \begin{equation}
        W_{\lambda}(\mu, \nu) = \min_{\gamma \in \Gamma(\mu, \nu)} \sum_{i=1}^{n} \sum_{j=1}^{m} \gamma_{ij} d(x_i, y_j) + \frac{1}{\lambda} \sum_{i=1}^{n} \sum_{j=1}^{m} \gamma_{ij} (\log \gamma_{ij} - 1).
    \end{equation}
    
    In FWI, we use the Wasserstein-Sinkhorn distance to measure the difference between observed data $d_{obs}(s, r, t)$ and synthetic $d_{cal}(\mathbf{m}; s, r, t)$ for each source-receiver pair $(s, r)$. The objective function based on the Wasserstein-Sinkhorn distance is defined as (Engquist et al., 2016; M{\'e}tivier et al., 2016; Y. Yang et al., 2018; Chizat et a., 2020):
    \begin{equation}
        \mathcal{J}_{WS}(\mathbf{m}) = \sum_{s} \sum_{r} W_{\lambda}(d_{obs}(s, r, t), d_{cal}(\mathbf{m}; s, r, t)),
    \end{equation}
    where $W_{\lambda}(d_{obs}(s, r, t), d_{cal}(\mathbf{m}; s, r, t))$ represents the Wasserstein-Sinkhorn distance between the observed and simulated data. This objective function provides a robust and flexible measurement of misfits.

\clearpage

%
%

\noindent\textbf{Text S2: Optimization methods}

\vspace{10pt}
\noindent\textbf{(1) Root mean square propagation}

    The root mean square propagation optimizer (RMSProp) is used to update the model parameters $\mathbf{m}$ by adjusting the learning rate based on the moving average of the squared gradients. The update rules are (Graves, 2014):
    \begin{align}
        g_t &= \nabla_{\mathbf{m}} \mathcal{J}(\mathbf{m}_{t-1}), \\
        E[g_t^2] &= \gamma E[g_{t-1}^2] + (1 - \gamma) g_t^2, \\
        \mathbf{m}_{t} &= \mathbf{m}_{t-1} - \frac{\eta}{\sqrt{E[g_t^2] + \epsilon}} g_t,
    \end{align}
    where $\mathcal{J}$ represents the objective function, $\mathbf{m_t}$ denotes the model parameters at time step $t$. The term $E[g_t^2]$ is the exponentially weighted moving average of the squared gradients $g_t$, with the decay rate $\gamma$ controlling the influence of past gradient magnitudes. The learning rate $\eta$ is a scalar factor that controls the step size, and $\epsilon$ is a small constant included for numerical stability to prevent division by zero.

\vspace{15pt}
\noindent\textbf{(2) Adaptive gradient algorithm}

    The adaptive gradient optimizer (Adagrad) can adjust the learning rate for each model parameter based on the accumulation of historical gradients (Duchi et al., 2011):
    \begin{align}
        g_t &= \nabla_{\mathbf{m}} \mathcal{J}(\mathbf{m}_{t-1}), \\
        G_t &= G_{t-1} + g_t^2, \\
        \mathbf{m}_{t} &= \mathbf{m}_{t-1} - \frac{\eta}{\sqrt{G_t + \epsilon}} g_t,
    \end{align}
    where $G_t$ accumulates the sum of squared gradients for the model parameters up to time $t$, and $\epsilon$ ensures numerical stability.

\vspace{15pt}
\noindent\textbf{(3) Adaptive moment estimation}

    The adaptive moment estimation optimizer (Adam) estimates the first and second moments of the gradients to compute adaptive learning rates for model parameters $\mathbf{m}$. The update rules are as follows (Kingma \& Ba, 2017):
    \begin{align}
        v_t &= \beta_1 v_{t-1} + (1 - \beta_1) g_t, \\
        s_t &= \beta_2 s_{t-1} + (1 - \beta_2) g_t^2, \\
        \hat{v}_t &= \frac{v_t}{1 - \beta_1^t}, \\
        \hat{s}_t &= \frac{s_t}{1 - \beta_2^t}, \\
        \mathbf{m}_{t} &= \mathbf{m}_{t-1} - \eta \frac{\hat{v}_t}{\sqrt{\hat{s}_t} + \epsilon},
    \end{align}
    where $v_t$ and $s_t$ represent the estimates of the first and second moments of the gradients, $\beta_1$ and $\beta_2$ are the decay rates, and $\epsilon$ is a small constant for numerical stability.
    
\vspace{10pt}
\noindent\textbf{(4) Adam with weight decay}

    Adam with weight decay (AdamW) uses decaying weights to regularize the model parameters. The update rules are (Loshchilov \& Hutter, 2019):
    \begin{align}
        v_t &= \beta_1 v_{t-1} + (1 - \beta_1) g_t, \\
        s_t &= \beta_2 s_{t-1} + (1 - \beta_2) g_t^2, \\
        \hat{v}_t &= \frac{v_t}{1 - \beta_1^t}, \\
        \hat{s}_t &= \frac{s_t}{1 - \beta_2^t}, \\
        \mathbf{m}_t &= \mathbf{m}_{t-1} - \eta \frac{\hat{v}_t}{\sqrt{\hat{s}_t} + \epsilon} + \lambda \eta \mathbf{m}_{t-1},
    \end{align}
    where $\lambda$ is the weight decay coefficient.

\vspace{10pt}
\noindent\textbf{(5) Nesterov-accelerated adaptive moment estimation}

    Nesterov-accelerated Adam (NAdam) integrates the Nesterov momentum to enhance convergence speed during the optimization process. The update rules are (Dozat, 2016):
    \begin{align}
        v_t &= \beta_1 v_{t-1} + (1 - \beta_1) g_t, \\
        s_t &= \beta_2 s_{t-1} + (1 - \beta_2) g_t^2, \\
        \hat{v}_t &= \frac{v_t}{1 - \beta_1^t}, \hat{s}_t = \frac{s_t}{1 - \beta_2^t}, \\
        \mathbf{m}_t &= \mathbf{m}_{t-1} - \frac{\eta}{{\sqrt{\hat{s}_t} + \epsilon}} (\beta_1 \hat{v}_t +\frac{1 - \beta_1}{1 - \beta_1^k}g_t).
    \end{align}

\vspace{15pt}
\noindent\textbf{(6) Rectified Adam}

    Rectified Adam (RAdam) introduces a rectification mechanism into Adam to stabilize the variance of adaptive learning rates, which is particularly useful in early optimization stages. The update rules are (L. Liu et al., 2019):
    \begin{align}
        v_{t} &= \beta_1 v_{t-1} + (1 - \beta_1) g_t, \\
        s_{t} &= \beta_2 s_{t-1} + (1 - \beta_2) g_t^2, \\
        \hat{v}_t &= \frac{v_t}{1 - \beta_1^t}, \\
        \hat{s}_t &= \frac{s_t}{1 - \beta_2^t}, \\
        \rho_\infty &= \frac{2}{1 - \beta_2} - 1, \\
        \rho_t &= \rho_\infty - \frac{2t\beta_2^t}{1 - \beta_2^t}, \\
        r_t &= \sqrt{\frac{(\rho_t - 4)(\rho_t - 2)\rho_\infty}{(\rho_\infty - 4)(\rho_\infty - 2)\rho_t}}, \\
        \textbf{m}_{t} &= \textbf{m}_{t-1} - \eta \frac{\hat{v}_t}{\sqrt{\hat{s}_t} + \epsilon} r_t.
    \end{align}

\clearpage

%
%
\vspace{15pt}
\begin{table}[!h]
\caption{Network architectures for the CNN-based neural networks for model reparameterization with varying numbers of CNN blocks. \\}
\centering
\begin{tabular}{|c|c|l|}
\hline
\multicolumn{1}{|c|}{\textbf{Model}} & \multicolumn{1}{c|}{\textbf{NN layer}} & \multicolumn{1}{c|}{\textbf{Network Architecture}} \\ \hline
\multirow{5}{*}{2 CNNs} & Input  & Random latent vector (100) \\
                               & Layer 1  & Fully-connected layer (4) + Leaky ReLu (0.1) + Reshape \\
                               & Layer 2  & 2x2 Upsampling + 4x4 Conv layer (4) + Leaky ReLU (0.1) + Dropout \\
                               & Layer 3  & 2x2 Upsampling + 4x4 Conv layer (32) + Leaky ReLU (0.1) + Dropout \\
                               & Output   & 4x4 Conv layer (1) + Tanh \\ 
                               \hline

\multirow{6}{*}{3 CNNs} & Input  & Random latent vector (100) \\
                               & Layer 1  & Fully-connected layer (16) + Leaky ReLu (0.1) + Reshape \\
                               & Layer 2  & 2x2 Upsampling + 4x4 Conv layer (16) + Leaky ReLU (0.1) + Dropout \\
                               & Layer 3  & 2x2 Upsampling + 4x4 Conv layer (32) + Leaky ReLU (0.1) + Dropout \\
                               & Layer 4  & 2x2 Upsampling + 4x4 Conv layer (16) + Leaky ReLU (0.1) + Dropout \\
                               & Output   & 4x4 Conv layer (1) + Tanh \\ \hline

\multirow{7}{*}{4 CNNs} & Input  & Random latent vector (100) \\
                               & Layer 1  & Fully-connected layer (64) + Leaky ReLu (0.1) + Reshape \\
                               & Layer 2  & 2x2 Upsampling + 4x4 Conv layer (64) + Leaky ReLU (0.1) + Dropout \\
                               & Layer 3  & 2x2 Upsampling + 4x4 Conv layer (32) + Leaky ReLU (0.1) + Dropout \\
                               & Layer 4  & 2x2 Upsampling + 4x4 Conv layer (16) + Leaky ReLU (0.1) + Dropout \\
                               & Layer 5  & 2x2 Upsampling + 4x4 Conv layer (16) + Leaky ReLU (0.1) + Dropout \\
                               & Output   & 4x4 Conv layer (1) + Tanh \\ \hline

\multirow{8}{*}{5 CNNs} & Input  & Random latent vector (100) \\
                               & Layer 1  & Fully-connected layer (256) + Leaky ReLu (0.1) + Reshape \\
                               & Layer 2  & 2x2 Upsampling + 4x4 Conv layer (256) + Leaky ReLU (0.1) + Dropout \\
                               & Layer 3  & 2x2 Upsampling + 4x4 Conv layer (32) + Leaky ReLU (0.1) + Dropout \\
                               & Layer 4  & 2x2 Upsampling + 4x4 Conv layer (32) + Leaky ReLU (0.1) + Dropout \\
                               & Layer 5  & 2x2 Upsampling + 4x4 Conv layer (32) + Leaky ReLU (0.1) + Dropout \\
                               & Layer 6  & 2x2 Upsampling + 4x4 Conv layer (16) + Leaky ReLU (0.1) + Dropout \\
                               & Output   & 4x4 Conv layer (1) + Tanh \\ \hline
\end{tabular}
\end{table}

%
%

\begin{figure}
  \centering
  \includegraphics[width=0.85\textwidth]{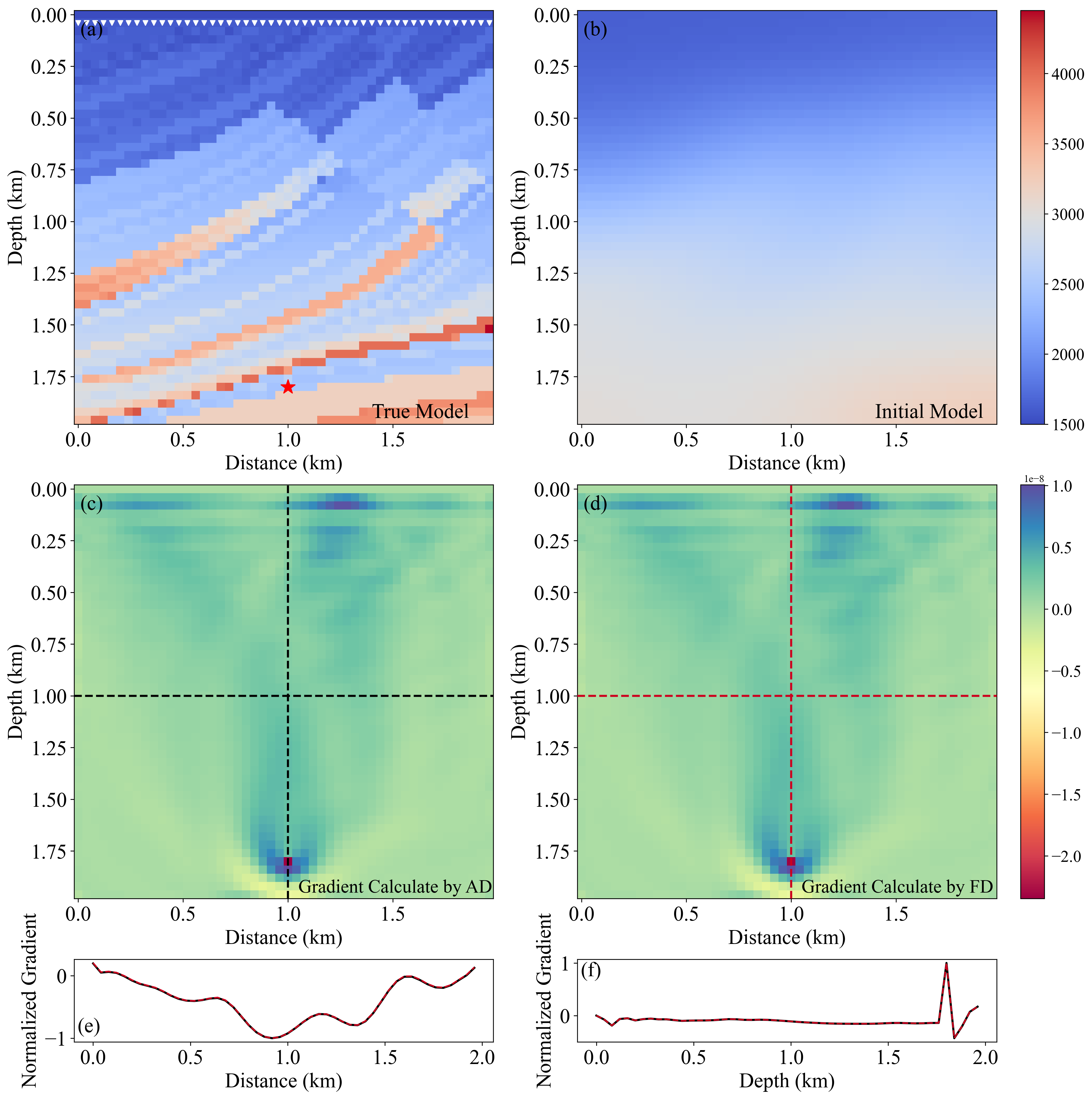}
  \caption{Comparison of gradients obtained through automatic differentiation (AD) and numerical differentiation computed explicitly with the central difference scheme (FD), the latter of which with a small $\Delta m$ is regarded as the ground truth in this case. (a) True velocity model ($v_p$), with the red star indicating the source and the white triangles indicating the receivers. (b) Initial velocity model. (c) Gradient calculated using AD for the single source and all receivers based on the initial model. (d) Gradient calculated using the FD method (ground truth). (e) Comparison of the gradients along the horizontal dashed lines in (c) and (d). (f) Comparison of gradients along the vertical lines. In (e) and (f), the black lines represent the gradients calculated by AD, and the red lines represent the ground-truth gradients by FD, which are essentially identical.}
  \label{FigureS1}
\end{figure}

\clearpage

\begin{figure}
  \centering
  \includegraphics[width=1.0\textwidth]{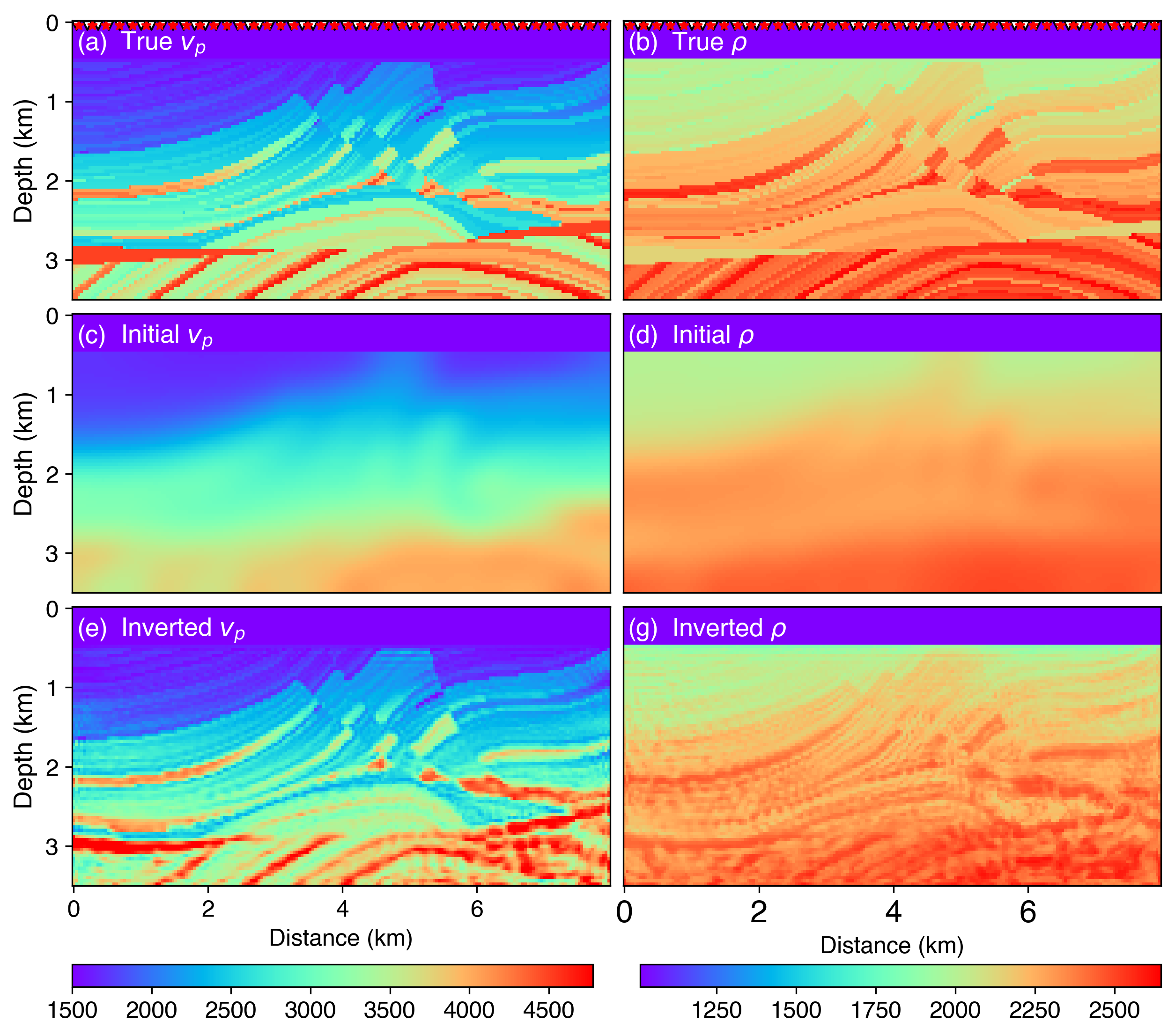}
  \caption{FWI test using the isotropic acoustic Marmousi2 model for simultaneous multi-parameters inversion for $v_p$ and $\rho$. (a)-(b) True models for $v_p$ and $\rho$, where the red stars indicate the source locations and the white triangles indicate the receiver positions. (c)-(d) Initial $v_p$ and $\rho$ models which are obtained by applying a 240 $\times$ 240 m Gaussian smoothing filter to the true models. (e)-(f) Simultaneously inverted $v_p$ and $\rho$ models after 500 iterations, where $\rho$ is updated once in every ten iterations to reduce inversion instability.}
  \label{FigureS2}
\end{figure}

\clearpage

\begin{figure}
    \centering
    \includegraphics[width=1.0\textwidth]{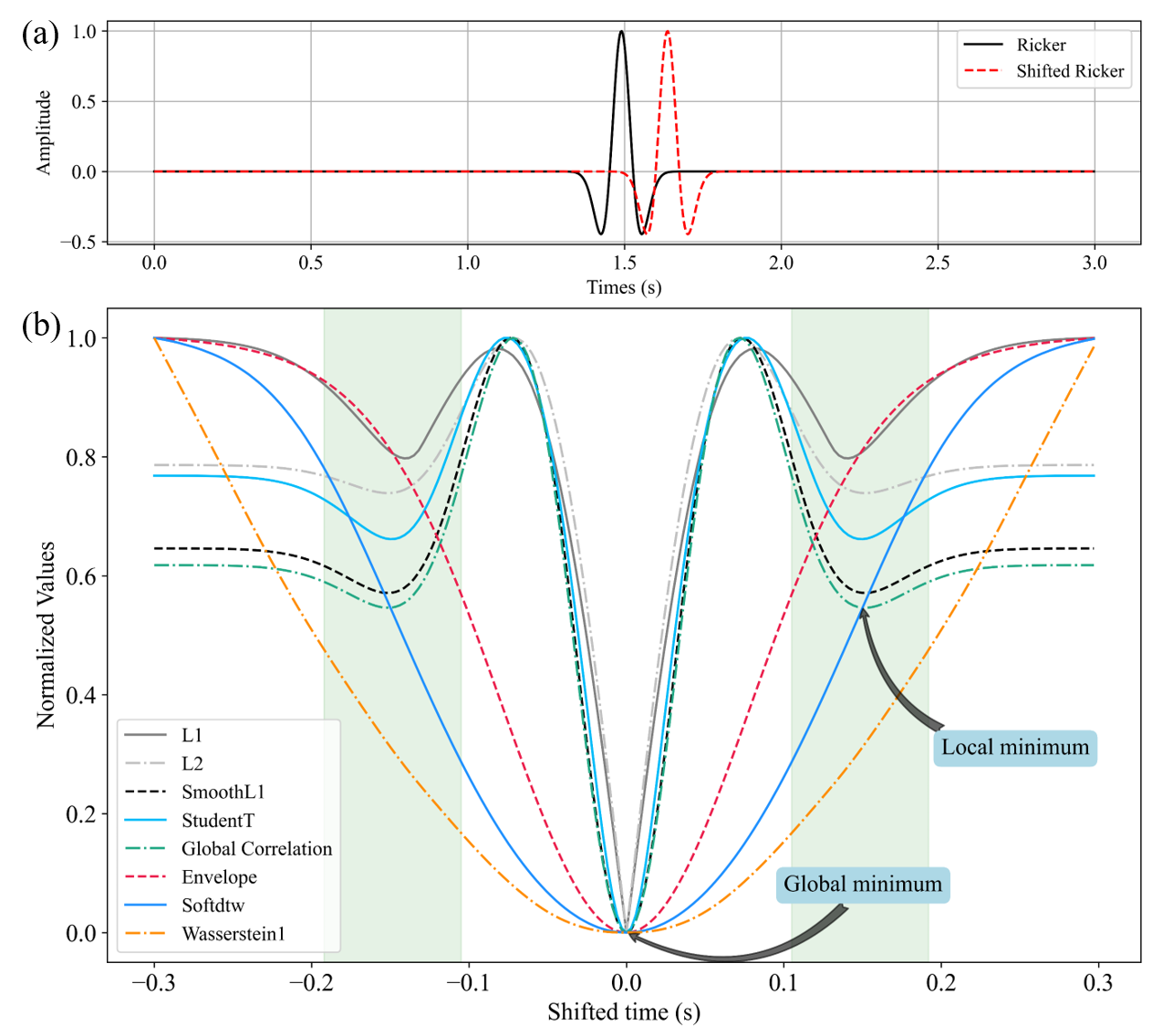}
    \caption{Comparison of various objective functions using a time-shifted Ricker wavelet with dominant frequency of 6 Hz. (a) Time-shifted Ricker wavelet (dotted red line) by 0.2 s compared to the original wavelet (solid black line). (b) Normalized variations for different objective functions with shifted time.}
    \label{FigureS3}
\end{figure}

\clearpage

\begin{figure}
  \centering
  \includegraphics[width=1.0\textwidth]{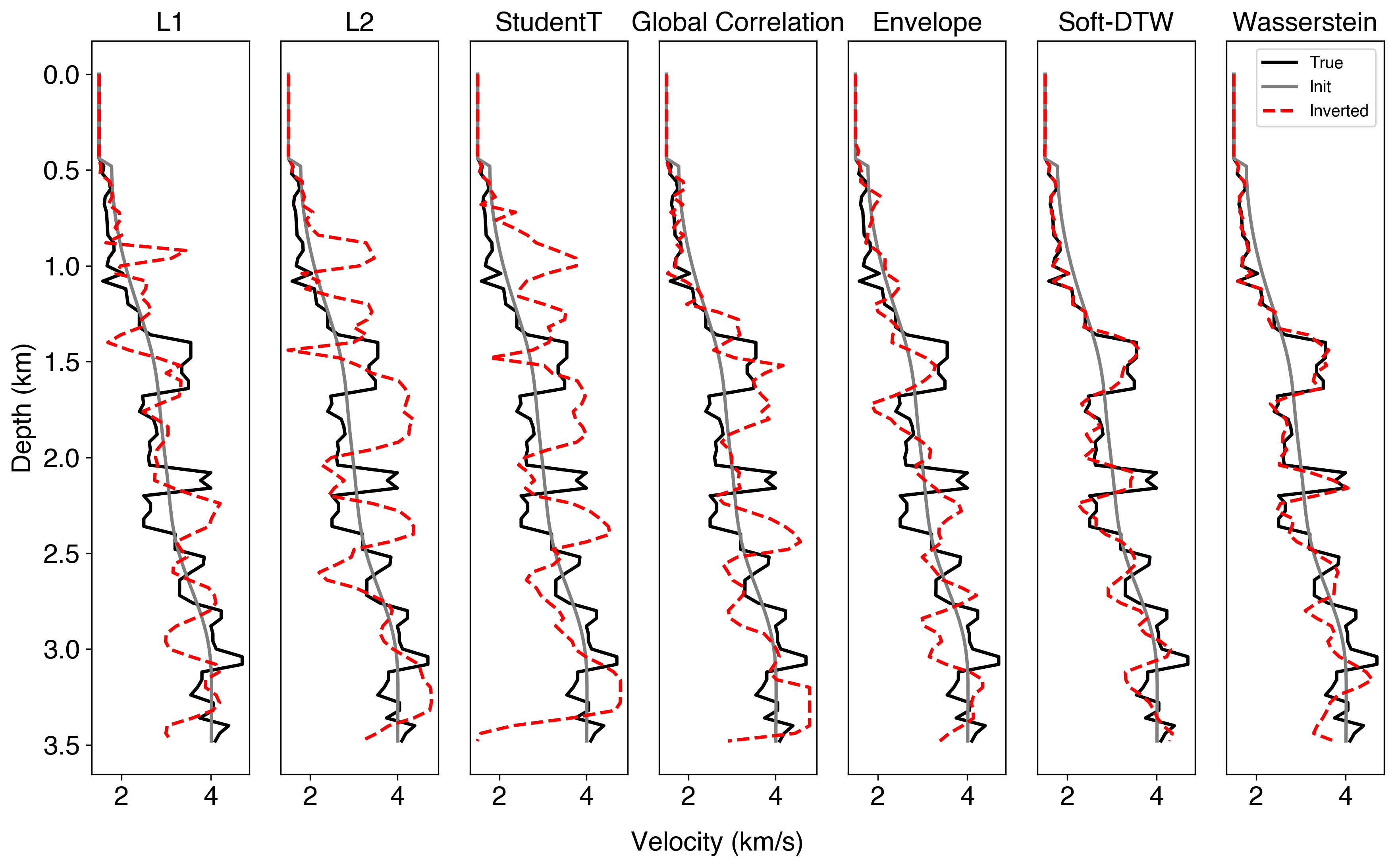}
  \caption{Comparison of profiles extracted from the models inverted with various objective functions at the lateral position of 5.5 km. The profiles presented here are from the models shown in Figure 6 in the main text.}
  \label{FigureS4}
\end{figure}

\clearpage

\begin{figure}
  \centering
  \includegraphics[width=1.0\textwidth]{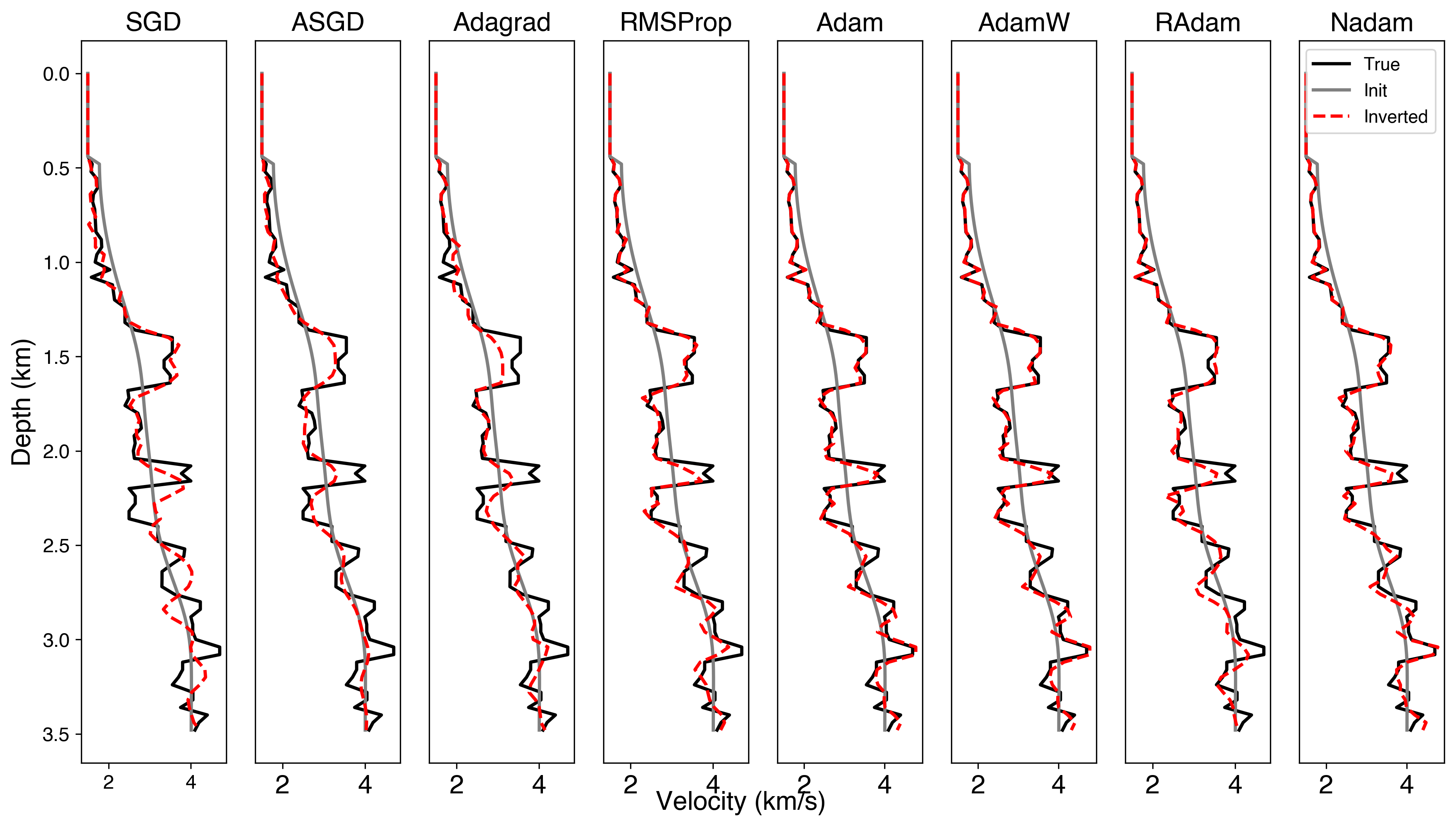}
  \caption{Comparison of profiles extracted from the models inverted with various optimization methods at the lateral position of 5.5 km. The profiles presented here are from the models shown in Figure 8 in the main text.}
  \label{FigureS5}
\end{figure}

\clearpage

\begin{figure}
  \centering
  \includegraphics[width=1.0\textwidth]{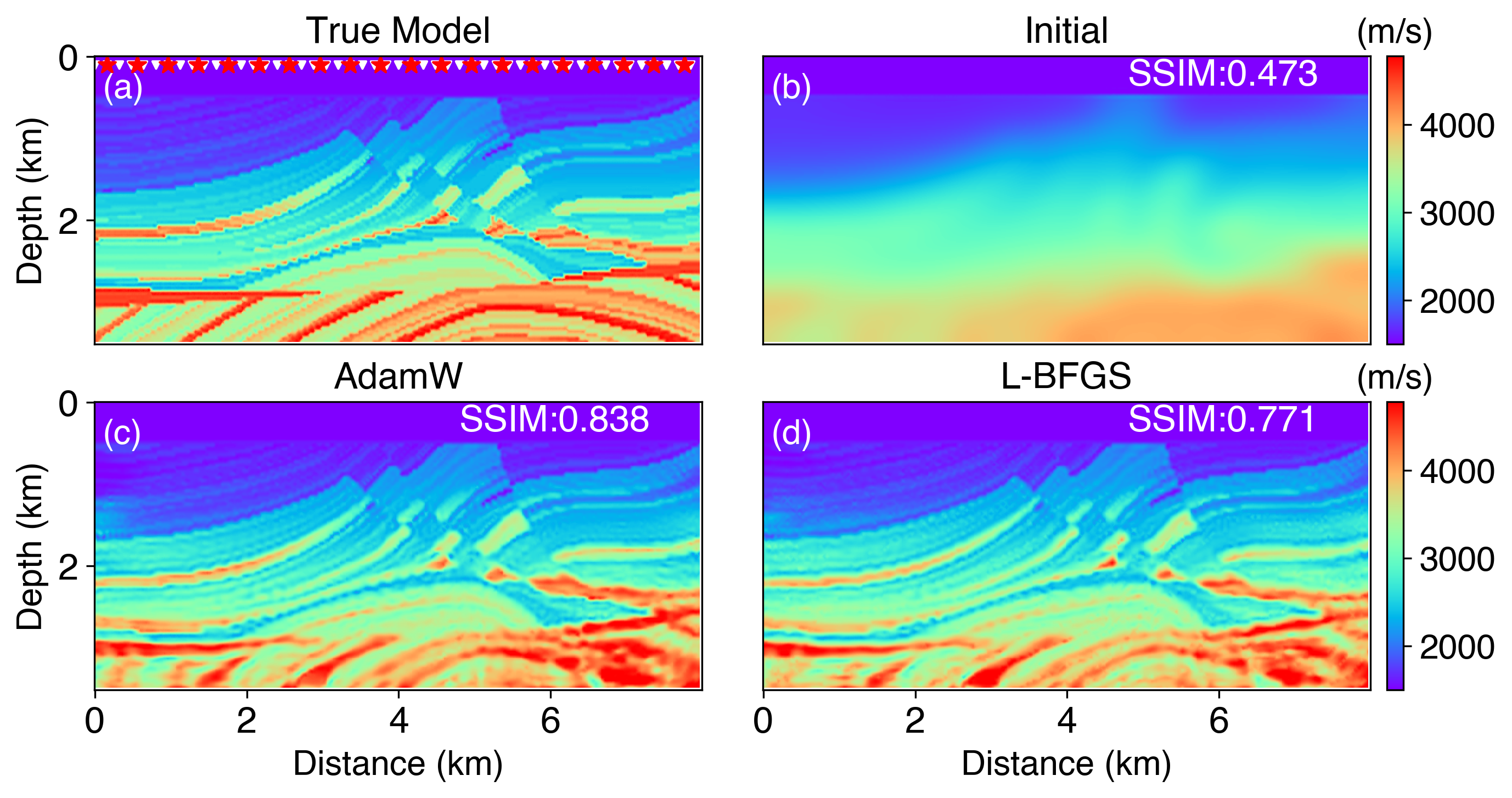}
  \caption{Comparison of the inversion results using the \textit{l}-BFGS and AdamW optimizers. (a) The true velocity model, with the red stars indicating the source locations and white triangles indicating the receiver positions. (b) The initial velocity model obtained by applying a 480 $\times$ 480 m Gaussian smoothing window to the true model. (c) Inversion result using the AdamW optimizer. (d) Inversion result using the \textit{l}-BFGS optimizer. The starting learning rate is 1, and the model is updated with 50 iterations with the \textit{l}-BFGS optimizer. In each iteration, a maximum of 25 line searches are conducted.}
  \label{FigureS6}
\end{figure}

\clearpage

\begin{figure}
  \centering
  \includegraphics[width=1.0\textwidth]{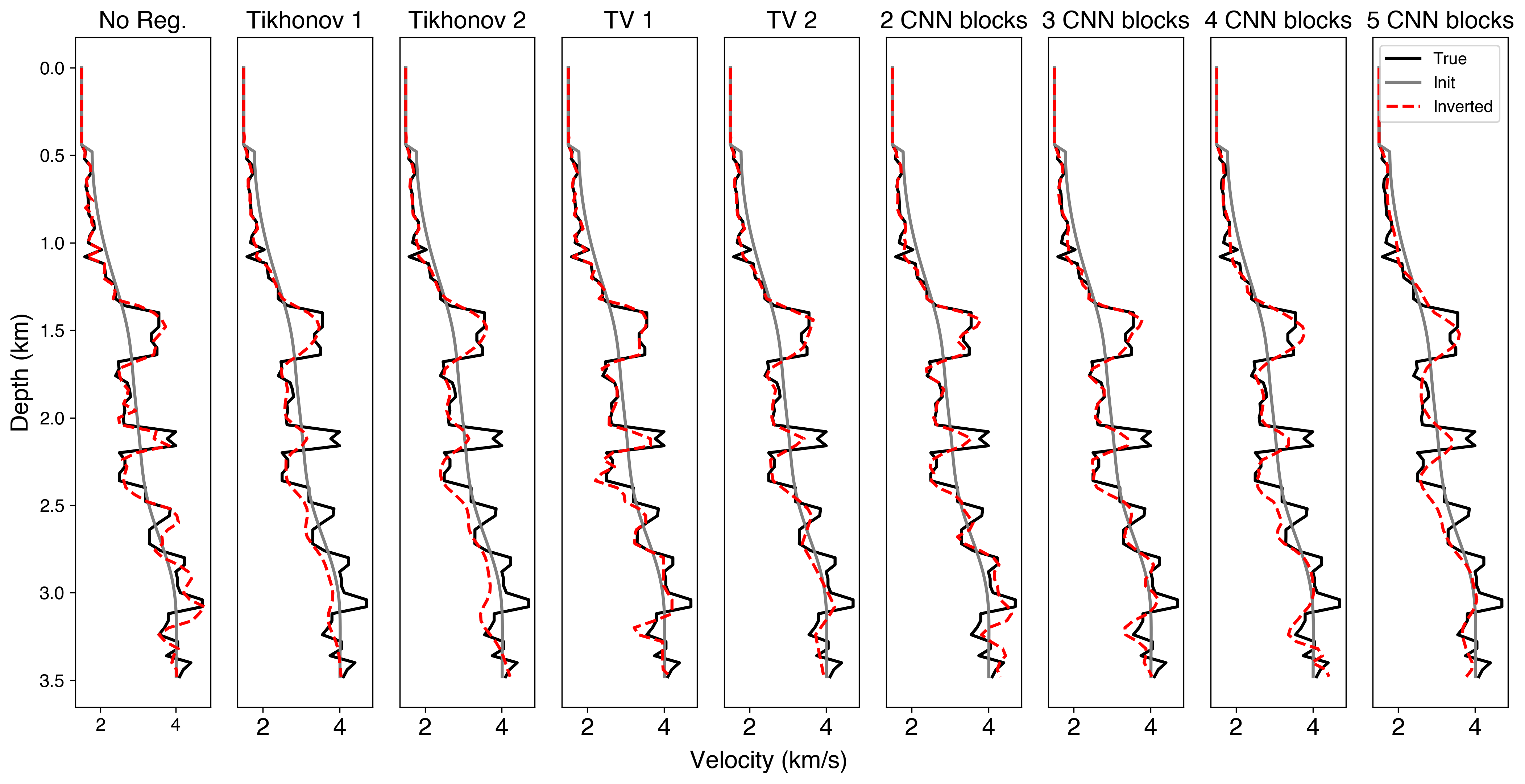}
  \caption{Comparison of profiles extracted from the models inverted with traditional regularization methods and CNN-based reparametrization methods at the lateral position of 5.5 km. The profiles presented here are from the models shown in Figures 10 and 12 in the main text.}
  \label{FigureS7}
\end{figure}